\documentclass{article}

\PassOptionsToPackage{numbers}{natbib}

\usepackage[preprint]{neurips_2025}

\usepackage[utf8]{inputenc} 
\usepackage[T1]{fontenc}    
\usepackage{hyperref}       
\usepackage{url}            
\usepackage{booktabs}       
\usepackage{amsfonts}       
\usepackage{nicefrac}       
\usepackage{microtype}      
\usepackage{xcolor}         

\usepackage{threeparttable}
\usepackage{multirow}
\usepackage{array}
\usepackage{float}
\usepackage{stfloats}
\usepackage{caption}
\usepackage{graphicx}
\usepackage{amsmath}
\usepackage{amssymb}
\usepackage{fix-cm}
\usepackage{mathrsfs} 
\usepackage{subcaption}
\usepackage{colortbl}
\usepackage{nicematrix}  
\usepackage{tikz}
\usepackage{wrapfig}

\title{UniCMs: A Unified Consistency Model For Efficient Multimodal Generation and Understanding}

\author{Chenkai Xu$^{1}$\thanks{Equal contribution.}\,\,, 
        Xu Wang$^{1}$\footnotemark[1]\,\,, 
        Zhenyi Liao$^{1}$, 
        Yishun Li$^{3}$, 
         Tianqi Hou$^{2}$, 
        Zhijie Deng$^{1}$\thanks{Corresponding author.} \\
        \textsuperscript{1}Shanghai Jiao Tong University\; \textsuperscript{2}Huawei \; \textsuperscript{3}Tongji University\\
        \texttt{\{132435xck,wangxu60,zhijied\}@sjtu.edu.cn} \\
}

\begin{document}

\maketitle

\begin{abstract}
Consistency models (CMs) have shown promise in the efficient generation of both image and text. 
This raises the natural question of whether we can learn a unified CM for efficient multimodal generation (e.g., text-to-image) and understanding (e.g., image-to-text). 
Intuitively, such a model could be acquired by applying the consistency distillation (CD) to existing unified multimodal models. 
However, the key challenge is establishing a unified denoising perspective for both image and text generation, which is essential for establishing the consistency mapping. 
To tackle this, at the representation level, we advocate for discrete tokens for both modalities to best preserve language modeling capabilities. 
Critically, instead of defining the text denoising trajectory via recent discrete diffusion language modeling principles, we specify it using the parallel decoding trace of an autoregressive language model, benefiting from the latter's superior performance in general text generation tasks. 
The denoising trajectory of image tokens adheres to standard discrete diffusion. 
We train our unified consistency models (UniCMs) on these combined multimodal trajectories simultaneously with a unified objective. 
We introduce a trajectory segmentation strategy to further improve the training convergence. 
Empirically, in text-to-image generation, UniCMs outperform SD3 on GenEval, Image Reward, and CLIP Score metrics, while requiring only approximately ${1}/{8}$ of the sampling time. 
Meanwhile, in image-to-text generation, 
UniCMs surpass Show-o on the MMMU benchmark while being $1.5 \times$ faster at long-sequence generating speed. 
The code is available at \url{https://github.com/zhijie-group/UniCMs}.

\end{abstract}

\begin{figure}[htbp]
    \centering
    \begin{minipage}{0.2\textwidth}
        \includegraphics[width=\linewidth]{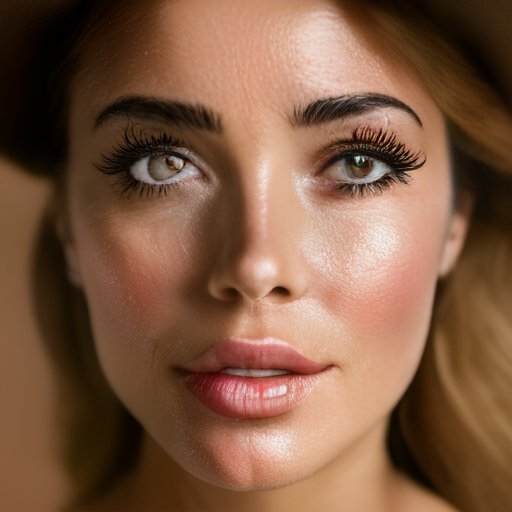}
    \end{minipage}%
    \begin{minipage}{0.2\textwidth}
        \includegraphics[width=\linewidth]{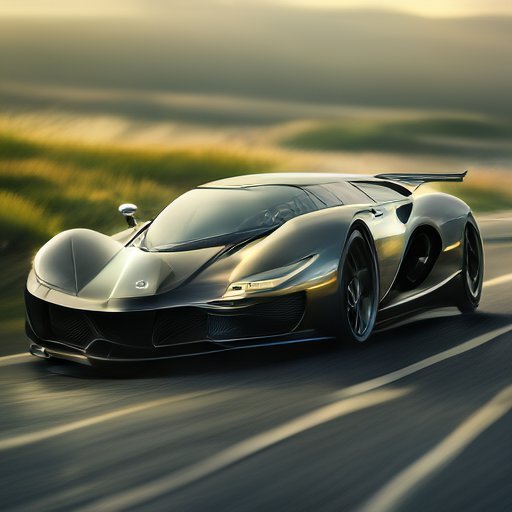}
    \end{minipage}%
    \begin{minipage}{0.2\textwidth}
        \includegraphics[width=\linewidth]{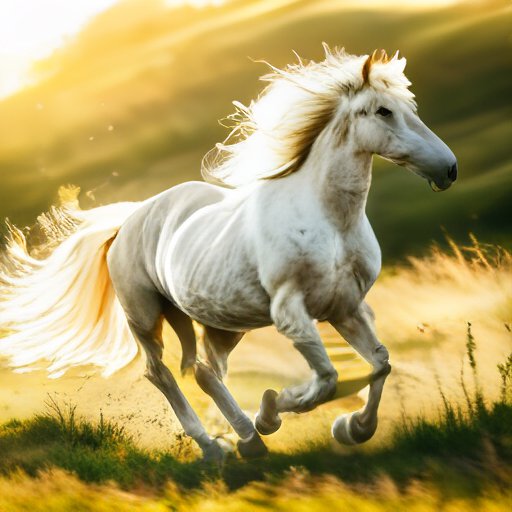}
    \end{minipage}%
    \begin{minipage}{0.2\textwidth}
        \includegraphics[width=\linewidth]{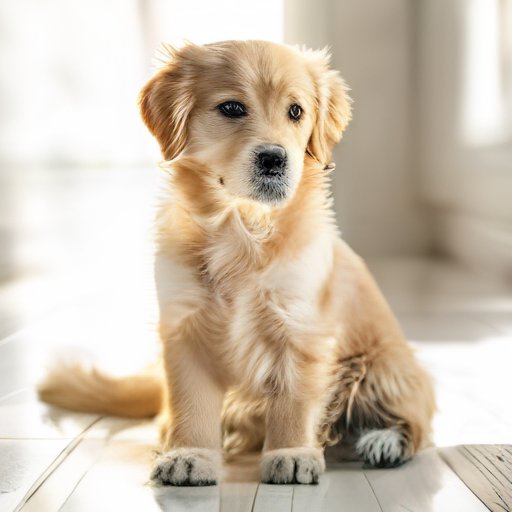}
    \end{minipage}%
    \begin{minipage}{0.2\textwidth}
        \includegraphics[width=\linewidth]{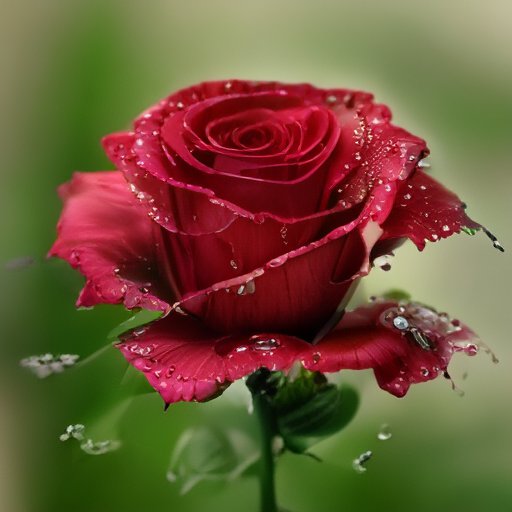}
    \end{minipage}%

    \vspace{-0.1cm}
    \begin{minipage}{0.2\textwidth}
        \includegraphics[width=\linewidth]{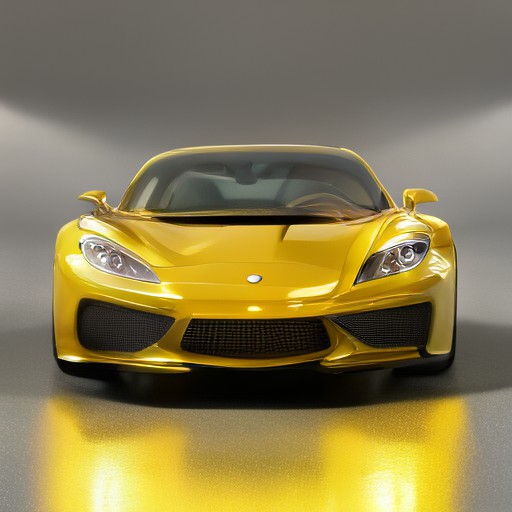}
    \end{minipage}%
    \begin{minipage}{0.2\textwidth}
        \includegraphics[width=\linewidth]{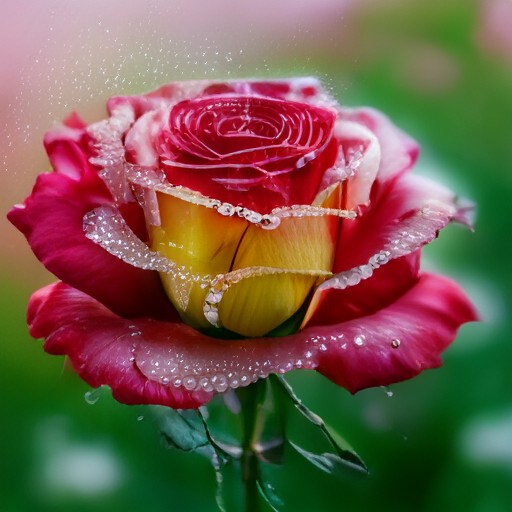}
    \end{minipage}%
    \begin{minipage}{0.2\textwidth}
        \includegraphics[width=\linewidth]{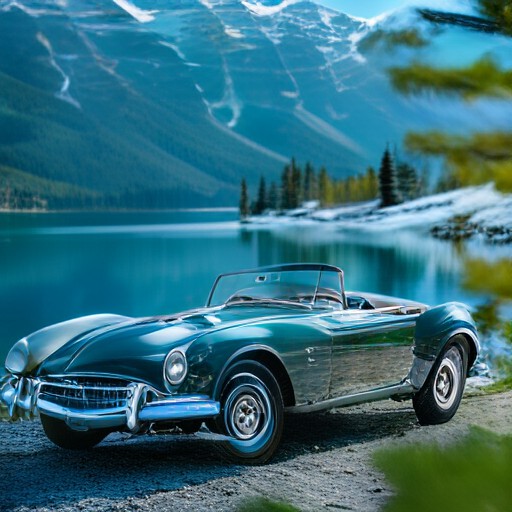}
    \end{minipage}%
    \begin{minipage}{0.2\textwidth}
        \includegraphics[width=\linewidth]{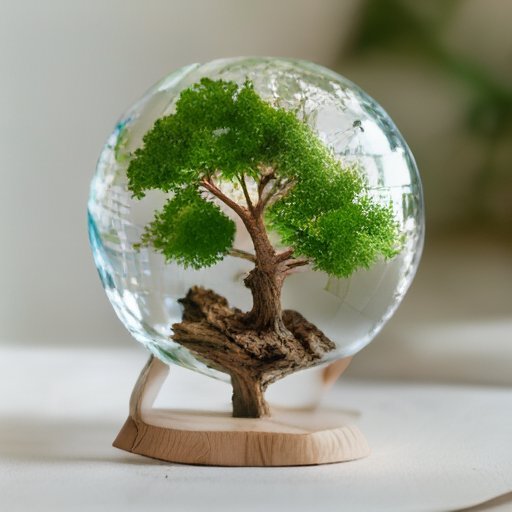}
    \end{minipage}%
    \begin{minipage}{0.2\textwidth}
        \includegraphics[width=\linewidth]{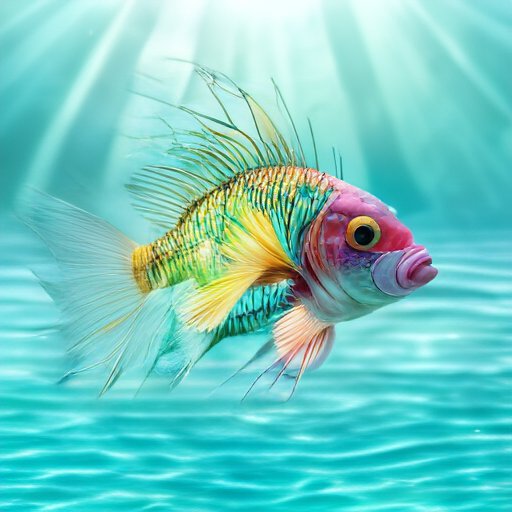}
    \end{minipage}%

    \vspace{-0.1cm}
    
    \begin{minipage}{0.2\textwidth}
        \includegraphics[width=\linewidth]{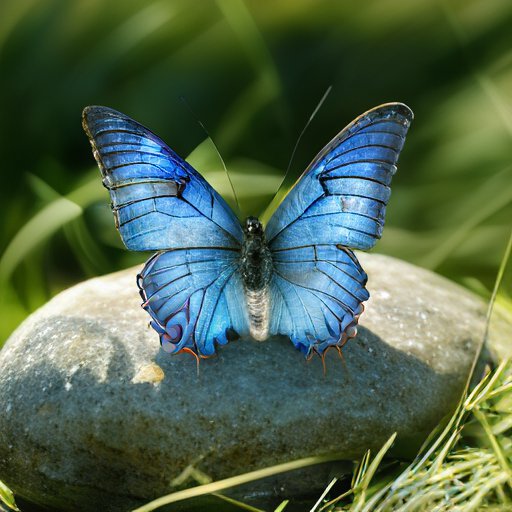}
    \end{minipage}%
    \begin{minipage}{0.2\textwidth}
        \includegraphics[width=\linewidth]{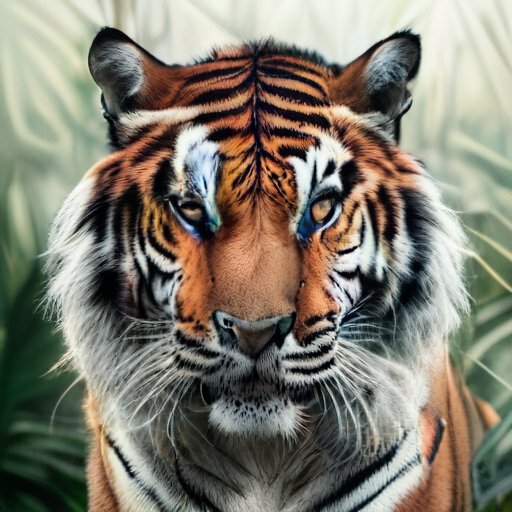}
    \end{minipage}%
    \begin{minipage}{0.2\textwidth}
        \includegraphics[width=\linewidth]{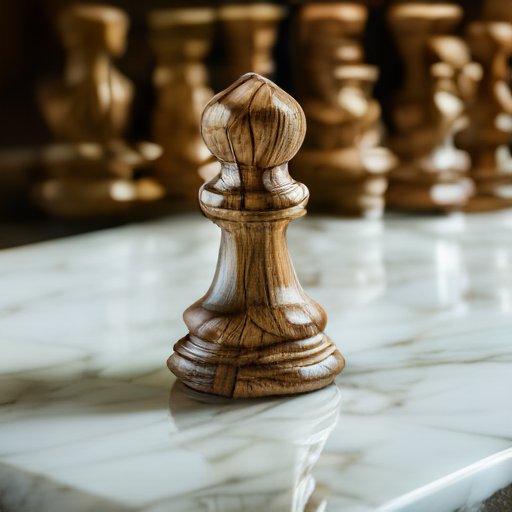}
    \end{minipage}%
    \begin{minipage}{0.2\textwidth}
        \includegraphics[width=\linewidth]{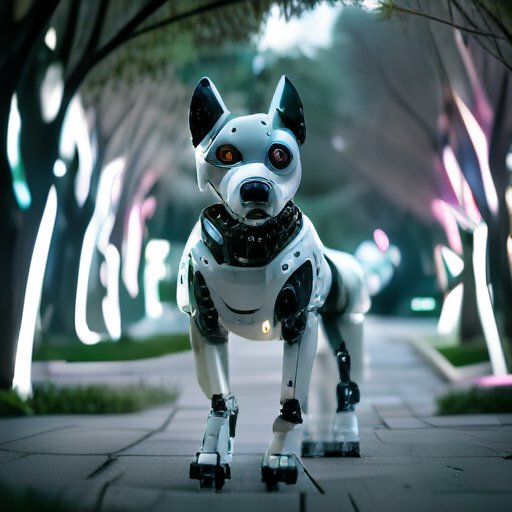}
    \end{minipage}%
    \begin{minipage}{0.2\textwidth}
        \includegraphics[width=\linewidth]{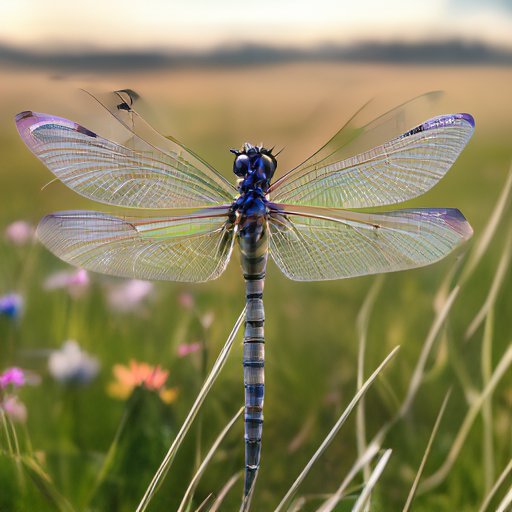}
    \end{minipage}%

    \vspace{-0.1cm}
    
    \begin{minipage}{0.2\textwidth}
        \includegraphics[width=\linewidth]{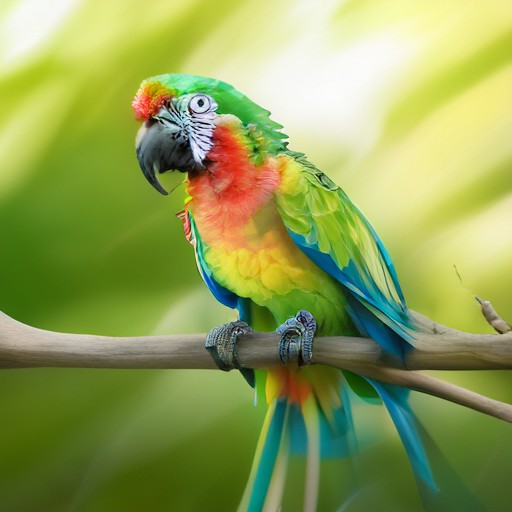}
    \end{minipage}%
    \begin{minipage}{0.2\textwidth}
        \includegraphics[width=\linewidth]{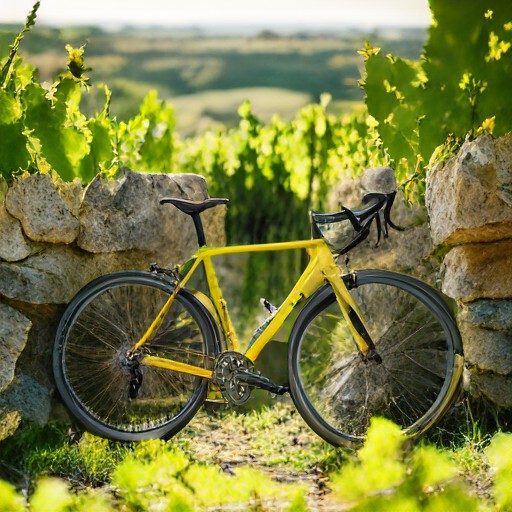}
    \end{minipage}%
    \begin{minipage}{0.2\textwidth}
        \includegraphics[width=\linewidth]{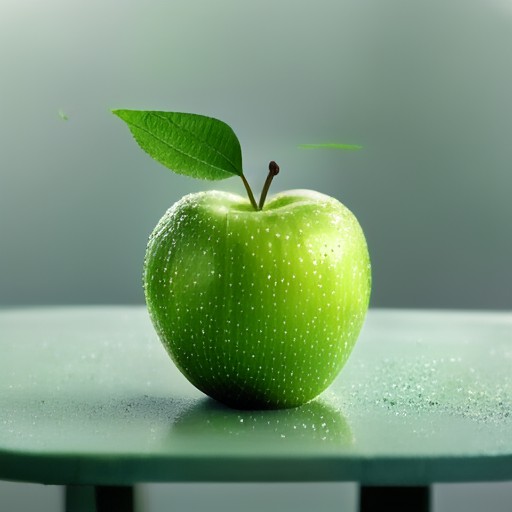}
    \end{minipage}%
    \begin{minipage}{0.2\textwidth}
        \includegraphics[width=\linewidth]{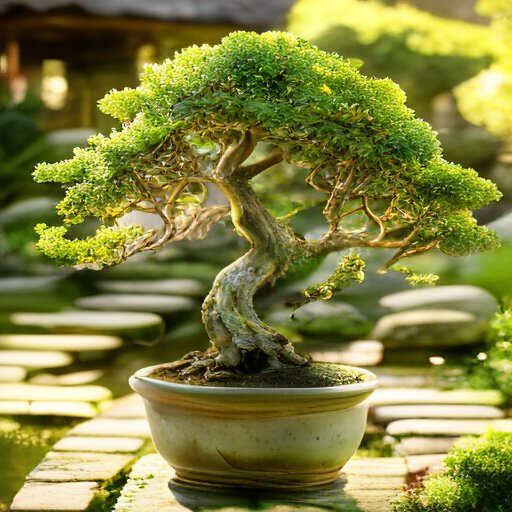}
    \end{minipage}%
    \begin{minipage}{0.2\textwidth}
        \includegraphics[width=\linewidth]{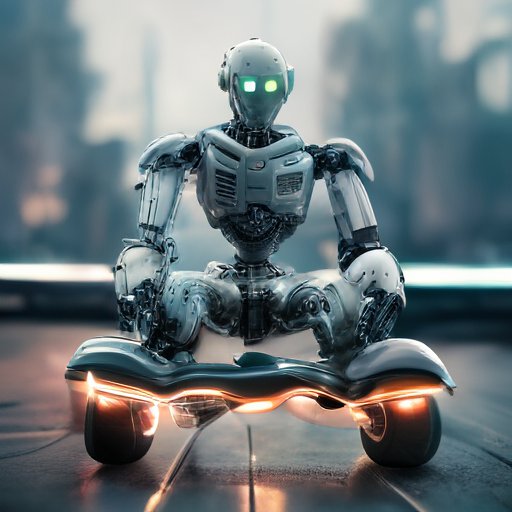}
    \end{minipage}%

    \caption{\textbf{512 $\times$ 512 images generated by UniCMs.} All images are generated by UniCMs in 4 sampling steps without reliance on classifier-free guidance~\cite{ho2021classifier}.}
    \label{fig:show}
    \vspace{-0.5cm}
\end{figure}


\section{Introduction}
\label{sec:intro}
Consistency models (CMs)~\cite{song2023consistency} have made significant achievements in efficient content generation across modalities. 
For image generation, CMs have revolutionized diffusion models, synthesizing high-fidelity images with few sampling steps~\cite{song2023consistency, luo2023latent, song2023improved, ren2024hyper, xie2024mlcm, wang2024phased}. 
Recently, 
CMs have been extended to text generation, 
realizing inference acceleration up to $3$ times~\cite{kou2024cllms}. 
Naturally, this raises an important question: \emph{can such advances in different modalities lead to a unified consistency model capable of efficiently understanding and generating cross-modal data?} 

Given the recent progress on unified multimodal generation and understanding models~\cite{team2405chameleon,zhou2024transfusion,wang2024emu3,xie2024show}, it is intuitive to apply consistency distillation (CD)~\cite{song2023consistency} to them to acquire unified consistency models. 
However, this cannot be implemented trivially due to a dilemma---the consistency mapping needs to be defined on a denoising-style generation trajectory, but how to establish a unified denoising perspective that encompasses both text and image generation remains an open challenge. 

This paper aims to address this. 
We first advocate for discrete tokenization for both modalities at the data representation level, which avoids degraded language modeling abilities. 
Thus, the core problem boils down to constructing a unified discrete denoising trajectory for the generation of both image and text tokens. 
For the former, we follow the typical masked diffusion paradigm (e.g., Muse~\cite{chang2023muse}, MaskGit~\cite{chang2022maskgit}, MagVit~\cite{yu2023magvit}, and Show-o~\cite{xie2024show}). 
For the latter, we suggest specifying the denoising trajectory with the parallel decoding trace of an autoregressive (AR) language generation process, given the success of consistency LLMs (CLLMs)~\cite{kou2024cllms}. 
We bypass the recent discrete diffusion language models~\cite{nie2025large, dream2025,sun2022score,nie2024scaling,gong2024scaling} because of their slightly inferior performance and limited application in processing multimodal inputs compared to AR ones. 





With such multimodal trajectories, we train the unified consistency models (UniCMs) using a unified objective. 
Specifically, UniCMs 
are pushed to consistently map any point on the trajectory to the same endpoint to enable fast-forward generation. 
We introduce a trajectory segmentation strategy~\cite{heek2024multistep,zheng2024trajectory,xie2024mlcm} in which distillation is applied to each segment of the complete generation trajectory to improve convergence.
We also design regularizations to ensure the training stability.
Conceptually, our approach constitutes an empirical generalization of the original CMs~\cite{song2023consistency} to discrete denoising trajectories and establishes a cross-modal extension of CLLMs.


Given that Show-o~\cite{xie2024show} can perform AR generation for text tokens and mask diffusion generation for image tokens, we opt to leverage it to collect text-to-image denoising trajectories on COCO 2017~\cite{lin2014microsoft} and image-to-text ones on LLaVA instruction tuning dataset~\cite{liu2024visual}. 
We then initialize UniCMs with Show-o and perform fine-tuning on such trajectories.
This training lasts for 36 hours on 8 A100-40GB GPUs. 
For text-to-image generation, 
UniCMs outperform SD3~\cite{esser2024scaling} on GenEval~\cite{ghosh2023genevalobjectfocusedframeworkevaluating}, Image Reward (IR)~\cite{li2024eagle}, and CLIP Score (CS)~\cite{hessel2022clipscorereferencefreeevaluationmetric}, while requiring only approximately ${1}/{8}$ of time cost. 
For image-to-text generation, UniCMs 
surpass Show-o on the MMMU~\cite{yue2024mmmu} benchmark while being approximately $1.5\times$ faster on the captioning tasks like 
NoCaps~\cite{agrawal2019nocaps}. 
\section{Related Work}
\label{sec:related}

\noindent \textbf{Unified Models.}
Early generative models often specialized in either text-conditioned image generation~\cite{rombach2022high, podellsdxl, song2020score, chen2023pixart, chen2024pixart, li2024hunyuan, yang2024cogvideox, sun2024autoregressive} or vision language understanding~\cite{liu2024llava, lin2024moe, liu2024visual, liu2024improved, li2024llava, zhu2024llava, bai2023qwen, ye2024mplug, zhu2023minigpt}, typically handling only one direction of multimodal interaction. 
To overcome this limitation, unified multimodal models~\cite{wu2023next, zhao2024monoformer, chern2024anole, dong2023dreamllm, wu2024janus} have emerged that aim to handle both image and text tasks simultaneously. 
For instance, Chameleon~\cite{team2023gemini} and Emu3~\cite{wang2024emu3} autoregressively generate both text and image tokens, 
while Transfusion~\cite{zhou2024transfusion} combines the autoregressive and continuous diffusion generation methods to handle different tasks. 
Similar to Transfusion, Show-o~\cite{xie2024show} also applies the autoregressive text generation but adopts the discrete diffusion methods in image generation process. 
While these unified models signify a major step towards versatile multimodal models, their reliance on iterative generation often leads to substantial computational overhead and slow inference.

\noindent \textbf{Consistency Models (CMs).} 
CMs have attracted significant attention due to their ability to generate high-quality outputs efficiently. 
Initially proposed in the context of continuous diffusion models~\cite{song2023consistency, luo2023latent}, CMs introduce the notion of trajectory consistency: they are trained to map any two points along the same sampling trajectory to a common endpoint~\cite{song2023consistency}.
This self-consistency property allows the model to bypass intermediate steps and directly predict the trajectory's endpoint, facilitating high-quality generation in significantly fewer steps—potentially even a single step. 
Subsequent works are built upon this foundation, introducing multi-step consistency models that segment the trajectory and apply consistency objectives within each segment~\cite{zheng2024trajectory, heek2024multistep, xie2024mlcm, wang2024phased}.
While most research has focused on continuous domains, the consistency principle has also been explored for discrete diffusion models~\cite{hayakawa2024distillation}, though with noted limitations in efficiency gains. 
The consistency distillation paradigm has also been adapted to enhance the efficiency of large language models (LLMs) by applying similar principles to accelerate iterative text generation~\cite{kou2024cllms}. 
However, these efforts have largely concentrated on continuous diffusion models, primarily for image generation, or on purely text-based models addressing single tasks. 
To date, unified consistency models remain largely unexplored.

\section{Method}
\label{sec:method}
This section presents unified consistency models (UniCMs) for efficient multimodal generation and understanding. 
We first review existing approaches on unified models and then provide insights on how to establish a unified denoising trajectory for learning UniCMs. 
We also elaborate on the unified CD loss as well as a suite of strategies to improve the model training. 
\subsection{Preliminary: Unified Models for Multimodal Generation and Understanding}

Unified multimodal modeling aims to process both textual and visual modalities within a compact model for joint generation~\cite{team2023gemini,wang2024emu3,team2405chameleon}. 
Typically, the architecture includes a transformer backbone, an encoder and decoder for images, and a text tokenizer. 
The image encoder converts an input image into patch-wise tokens $\mathbf{u}=\{u_1,\dots,u_m\}$, where 
$m$ is the number of patches and $u_i$ can be continuous vectors or discrete indices derived from vector quantization~\cite{van2017neural}. 
The text tokenizer encodes text into $n$ discrete tokens $\mathbf{v}=\{v_1,\dots,v_n\}$. 
The unified model then characterizes the text-to-image (T2I) and image-to-text (i.e., multimodal understanding, MMU) relationships simultaneously with the shared transformer backbone.  
In particular, the backbone predicts image and text tokens, which are then decoded by the image decoder and detokenized, respectively, to obtain images and text.

Unified models typically generate text tokens $\mathbf{v}$ autoregressively, a consequence of language's discrete and sequential nature. Formally, the learning objective is Next Token Prediction (NTP):
\begin{equation}\small\label{eq:ntp}
\mathcal{L}_{\text{NTP}} := \sum_i \log p_\theta(v_i | v_1, \cdots , v_{i-1}, \mathbf{u}),
\end{equation}
where $\theta$ denotes learnable parameters and $p_\theta$ refers to model likelihood. 

Based on how $\mathbf{u}$ are produced, existing approaches can be categorized into three main classes:
\begin{itemize}
    \item Autoregressive generation~\cite{sun2024autoregressive,ma2024star} of $\mathbf{u}$, where $\mathbf{u}$ are discrete, as seen in models like Emu3~\cite{wang2024emu3}, Chameleon~\cite{team2405chameleon}, LWM~\cite{liu2024world}, etc.
    \item Discrete diffusion~\cite{chang2022maskgit,chang2023muse,yu2023magvit} generation of $\mathbf{u}$, which also relies on discrete $\mathbf{u}$ and is known as mask diffusion, exemplified by Show-o~\cite{xie2024show}.
    \item Gaussian diffusion~\cite{ho2020denoising,song2020score,luo2023latent} generation of $\mathbf{u}$, where $\mathbf{u}$ are continuous vectors, as demonstrated in Transfusion~\cite{zhou2024transfusion}.
\end{itemize}



Despite the promise of multimodal generation, unified models can be slow in generation speed, particularly in the T2I generation scenario. 
For example, Emu3 requires over one minute to generate an image of $512 \times 512$ resolution on an NVIDIA 4090 GPU, due to the lengthy nature of the image tokens (e.g., 4096). 
Models like Show-o and Transfusion can alleviate such an issue thanks to the diffusion-based modeling, but their speed still lags significantly behind specialized T2I generators~\cite{ren2024hyper,sauer2024adversarial}.
On the other hand, the image-to-text generation also requires acceleration because the output tokens can be numerous in some cases, e.g., the image captioning task~\cite{flickrentitiesijcv,flickr30k} and multimodal chain-of-thought reasoning task~\cite{shen2025vlm,feng2025video}. 

\textbf{Efficient Unified Generation and Understanding by CMs.} 
CMs are known for efficient content generation, with applications in both image~\cite{song2023consistency,luo2023latent} and text~\cite{kou2024cllms} generation, which provides the opportunity for a unified efficient generation framework.
Basically, given a denoising trajectory collected from the sampling process, CMs map any two points along it to a common endpoint to enable fast-forward generation. 
Thus, to learn unified CMs, we should identify a unified denoising trajectory for the two modalities. 
When using discrete image tokens to align with the discrete text ones for unified modeling, the core problem becomes identifying a unified discrete denoising trajectory.

\setcounter{figure}{1}
\begin{figure*}[t] 
    \centering
    \includegraphics[width=\textwidth,keepaspectratio]{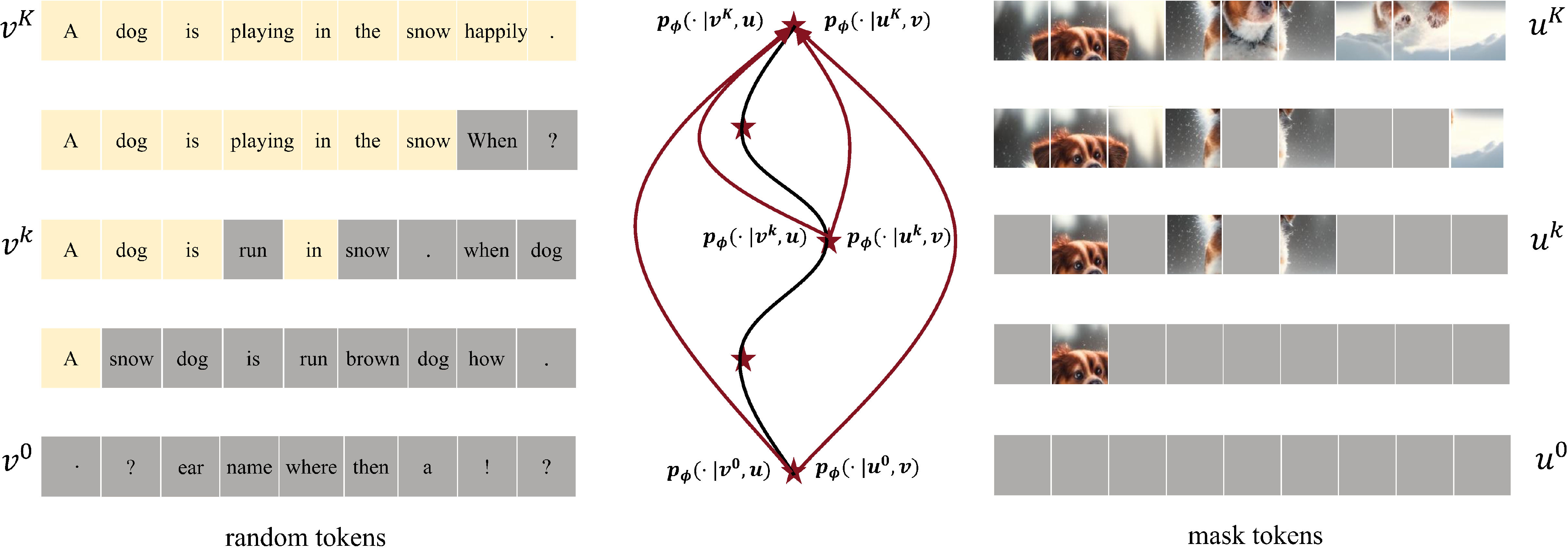}
    \caption{\textbf{Illustration of the unified denoising perspective of text and image generation.} 
    As shown, the trajectories both display a denoising pattern. 
    The black line denotes the unified abstraction of the multimodal trajectory, and the red lines illustrate the objective of UniCMs---to map an arbitrary point on the sampling trajectory to the same endpoint for both text and image generation. 
    Note that we omit the trajectory segmentation strategy in the training process for brevity.}
    \label{fig:trajectory_framework}
\end{figure*}

\subsection{A Unified Denoising Perspective for the Generation of Image and Text Tokens}




\noindent \textbf{Denoising Trajectory for Image.}
A natural approach to obtain a discrete denoising trajectory for image tokens $\mathbf{u}$ is through discrete diffusion modeling.
Typically, the process begins with a sequence of $m$ fully masked image tokens $\mathbf{u}^{0} := \{ u_1^{0}, \dots, u_m^{0}\}$, with the mask ratio progressively decreasing to 0 over $K$ iterative steps. 
Specifically, in the $k$-th step, given the sequence $\mathbf{u}^k$, let $M_k$ be the set of indices of masked tokens within $\mathbf{u}^k$.
The model first predicts the tokens for all masked positions $i \in M_k$ to get an intermediate sequence $\bar{\mathbf{u}}^{k+1}$ as follows:
\begin{equation}
    \bar{u}_i^{k+1} =
    \begin{cases}
        \arg \max_u p_\theta(u_i = u | \mathbf{u}^k, \mathbf{v}), & \text{if } i \in M_k \\
        u_i^{k}, & \text{if } i \notin M_k
    \end{cases}
\end{equation}
where $\mathbf{v}$ denotes the text condition and $p_\theta$ is abused to denote a T2I model that employs masked diffusion modeling on images (e.g., Show-o~\cite{xie2024show}, Muse~\cite{chang2023muse}, and Meissonic~\cite{bai2024meissonic}). 
Then, the model re-masks low-confidence generations in $\bar{\mathbf{u}}^{k+1}$ according to the schedule on mask ratio, yielding $\mathbf{u}^{k+1}$. 
The resultant trajectory $\{ \mathbf{u}^{0},\mathbf{u}^{1},\ldots,\mathbf{u}^K\}$ is visualized in Figure~\ref{fig:trajectory_framework}. 

\noindent \textbf{Denoising Trajectory for Text.}
To obtain text denoising trajectories, we consider two approaches: (1) leveraging recent discrete diffusion-based language generation methods~\cite{dream2025,nie2025large} or (2) utilizing the parallel decoding trajectories derived from an AR language generation process, as suggested by CLLMs~\cite{kou2024cllms}. 
Given the slightly inferior performance and limited application in processing multimodal inputs of diffusion language models compared to AR ones, we opt for the latter. 

Technically, starting from a sequence of $n$ randomly initialized text tokens, denoted as $\mathbf{v}^{0} := \{ v_1^{0}, \dots, v_n^{0}\}$, the parallel decoding process iteratively refines the token sequence until a fixed point. 
At $k$-th iteration, the refinement corresponds to simultaneously solving the following $n$ problems:
\begin{equation}\small
\begin{aligned}
    v_1^{k+1} &= \arg \max_v p_\theta(v | \mathbf{u}),\\
    v_2^{k+1} &= \arg \max_v p_\theta(v | v_1^{k}, \mathbf{u}),\\
    &...\\
    v_n^{k+1} &= \arg \max_v p_\theta(v | v_1^{k}, \dots, v_{n-1}^{k}, \mathbf{u}),
\end{aligned}
\end{equation}
where $p_\theta$ is abused for an image-to-text AR model. 
In fact, these problems can be solved simultaneously with only one forward pass using a causal attention mask, which takes roughly identical time as decoding one new token. 
Note that the greedy sampling strategy is used here. 
Abusing $K$ to denote the number of iterations to reach the fixed point $\mathbf{v}^{K}$, it is easy to see $K \leq n + 1$ because there is at least one token being correctly predicted in each iteration.\footnote{By correctness, we mean the generated tokens equal to those generated by regular AR decoding.}
Refer to Figure~\ref{fig:trajectory_framework} for a visualization of the sampling trajectory $\{ \mathbf{v}^{0}, \ldots, \mathbf{v}^{K}\}$, which displays a gradual denoising pattern.



\subsection{Training of UniCMs}
Based on the foregoing, text trajectories can be sourced from AR image-to-text models (like LLaVA~\cite{liu2023visual}, Qwen-VL-chat~\cite{bai2023qwen}, Show-o~\cite{xie2024show}), and image trajectories from mask diffusion T2I models (like Show-o~\cite{xie2024show}, Muse~\cite{chang2023muse}, and Meissonic~\cite{bai2024meissonic}). Given Show-o's ability to fulfill both roles, we favor it in our current work. Furthermore, this preference naturally extends to initializing UniCMs with Show-o's architecture and parameters when training on its trajectories, facilitating a smoother cold start. 
Letting $p_\phi$ denote the UniCMs to learn, we elaborate on the algorithmic details below. 


\noindent \textbf{Unified Training Objective.}
The consistency loss on image trajectories is:
\begin{equation}\small
\label{eq:lc_image}
\mathcal{L}^u_{c} = \mathbb{E}_{k \sim \mathcal{U}(0,K)}  d\left(p_{\phi^-}(\cdot|\mathbf{u}^K, \mathbf{v}), p_\phi(\cdot|\mathbf{u}^k, \mathbf{v})\right) ,
\end{equation}
where ${\phi^-}$ denotes stopping gradient backpropagation for stable training~\cite{song2023consistency} and $d$ indicates a divergence measure. 
For $\mathcal{L}^u_{c}$, $d$ aggregates the KL divergence between categorical prediction distributions over the masked image tokens. 
The consistency loss on text trajectories can be similarly defined:
\begin{equation}\small
\label{eq:lc_text} 
\mathcal{L}^v_{c} = \mathbb{E}_{k \sim \mathcal{U}(0,K)}  d\left(p_{\phi^-}(\cdot|\mathbf{u}, \mathbf{v}^{K}), p_\phi(\cdot|\mathbf{u}, \mathbf{v}^{k})\right) ,
\end{equation}
where $d$ aggregates over the positions where the two prediction distributions differ.
These losses, $\mathcal{L}^u_{c}$ and $\mathcal{L}^v_{c}$, are global consistency losses for image and text trajectory (mapping to their respective endpoints $\mathbf{u}^K$ and $\mathbf{v}^K$), empirically superior to local losses for discrete denoising trajectories~\cite{kou2024cllms}. 
Conceptually, our objective forms an empirical generalization of the original CMs defined on the ODE trajectories and a cross-modal extension of CLLMs~\cite{kou2024cllms}. 

\noindent \textbf{Trajectory Segmentation.}
We empirically ascertain that imposing long-range consistency may introduce unnecessary learning challenges, potentially impeding model convergence and ultimately limiting the model's inference efficiency.
Inspired by previous work~\cite{heek2024multistep,zheng2024trajectory,xie2024mlcm}, we design a segmentation strategy for the collected discrete multimodal sampled trajectories, enforcing consistency and regularization constraints in specific regions between points within a segment and segment endpoints.
More details about the trajectory segmentation can be found in Appendix~\ref{sec:segdetail}.
As the training proceeds, the trajectories of UniCMs may deviate significantly from the original collected multimodal trajectories. 
Thus, persisting in utilizing the original trajectory for distillation purposes could constrain the ultimate acceleration effect. 
We propose to regenerate multimodal denoising trajectories using the consistency model obtained in past stages. 
In this training stage, we also halve the number of segments of the trajectory to achieve better acceleration.
Doing so encourages the final UniCMs to learn consistency mapping over long distances.

\noindent \textbf{Regularization.}
Training UniCMs with only consistency loss in discrete multimodal denoising can lead to trivial convergence (e.g., identical outputs for varied inputs). To prevent this, we add regularizations for both modalities.
For text, $p_\phi$ must fit endpoint tokens $\mathbf{v}^K$ via an NTP objective. 
For images, 
we observe that the prediction logits of recovered image tokens contain rich information (e.g., easy-to-difficult hierarchies), so record them at each sampling step during trajectory collection (detailed in Appendix~\ref{sec:regdetail}, Figure~\ref{fig:reg}). 
Then, we use the logits as targets to regularize $p_\phi(\cdot|\mathbf{u}^k, \mathbf{v})$. 



We use $\mathcal{L}_{REG}^v$ and $\mathcal{L}_{REG}^u$ to represent these two regularizations respectively. 
The total loss is 
\begin{equation}
\mathcal{L} =\mathcal{L}_{c}^u + \alpha \mathcal{L}_{c}^v  +  \beta \mathcal{L}_{REG}^u + \gamma \mathcal{L}_{REG}^v,
\end{equation}
where $ \alpha $, $ \beta $ and $ \gamma $ are the trade-off coefficients to balance the different losses.

\begin{table*}[t]
    \centering
    \setlength{\tabcolsep}{4pt} 
    \definecolor{lightgray}{RGB}{230,230,230} 
    \begin{tabular}{llccccccc}
        \toprule
        Type & Model & Res. & Steps & GenEval $\uparrow$ & HPS $\uparrow$ & IR $\uparrow$ & CS $\uparrow$ & Time (s) $\downarrow$ \\
        \midrule
        \multirow{5}{*}{Gen. Only} & Emu3-Gen~\cite{wang2024emu3} & 512 & 4096 & 0.540 & - & - & - & 309.51 \\
        & SDXL~\cite{podell2023sdxl} & 1024 & 50 & 0.550 & 0.267 & 0.698 & 0.312 & 6.88 \\
        & SDXL-Turbo~\cite{sauer2024adversarial} & 512 & 1 & 0.551 & 0.273 & 0.759 & 0.315 &  0.27 \\
        & SD3~\cite{esser2024scaling} & 512 & 24 & 0.620 & 0.275 & 0.787 & 0.308 & 1.33 \\
        & Hyper-SD3~\cite{ren2024hyper} & 1024 & 4 & 0.458 & 0.266 & 0.649 & 0.308 & 1.19 \\
        \midrule
        \multirow{7}{*}{Und. \& Gen.} & Show-o~\cite{xie2024show} & 512 & 16 & \textbf{0.674} & \textbf{0.277} & \textbf{0.992} & \textbf{0.318} & 1.39 \\
        & Transfusion~\cite{zhou2024transfusion} & 256 & 250 & 0.630 & - & - & - & - \\
        & Chameleon~\cite{team2024chameleon}
        & 512 & 1024 & 0.430 & - & - & - & 19.24 \\
        & Orthus~\cite{team2024chameleon}
        & 512 & 1024 & 0.580 & - & - & - & 239.90 \\
        & \cellcolor{lightgray}{\multirow{3}{*}{\colorbox{lightgray}{UniCMs}}} & \cellcolor{lightgray}512 & \cellcolor{lightgray}8 & \cellcolor{lightgray}\underline{0.638} & \cellcolor{lightgray}\underline{0.273} & \cellcolor{lightgray}\underline{0.963} & \cellcolor{lightgray}\textbf{0.318} & \cellcolor{lightgray}0.33 \\
        & \cellcolor{lightgray}{UniCMs} & \cellcolor{lightgray}512 & \cellcolor{lightgray}4 & \cellcolor{lightgray}0.625 & \cellcolor{lightgray}0.269 & \cellcolor{lightgray}0.934 & \cellcolor{lightgray}\textbf{0.318} & \cellcolor{lightgray}\underline{0.17} \\
        & \cellcolor{lightgray}& \cellcolor{lightgray}512 & \cellcolor{lightgray}2 & \cellcolor{lightgray}0.557 & \cellcolor{lightgray}0.247 & \cellcolor{lightgray}0.680 & \cellcolor{lightgray}\underline{0.312} & \cellcolor{lightgray}\textbf{0.09} \\
        \bottomrule
    \end{tabular}
    \vspace{0.4cm}
    \caption{\textbf{Comparison of model performance for T2I task.} For the "Und. \& Gen." panel, best results are shown in \textbf{bold} and second best results are \underline{underlined}.}
    \label{tab:combined_comparison_full_with_more_cols_updated}
\end{table*}

\noindent \textbf{Sampling Strategy.}
We find that for the learned UniCMs with few sampling steps, there is significantly higher uncertainty in the prediction distribution of the mask tokens.
We empirically identify that incorporating the top-k sampling strategy, which is widely used in language models, can alleviate this issue, substantially improving the sampling quality in 2-4 steps (see Table~\ref{tab:topk_comparison}). 
\section{Experiments}
\label{sec:exp}
This section evaluates on T2I generation and MMU tasks to inspect the efficacy of UniCMs. 

\subsection{Implementation Details}
\label{sec:traindetails}

\noindent \textbf{Datasets.}
The captions from the training split of COCO 2017~\cite{lin2014microsoft} are used to generate text-to-image denoising trajectories. 
The LLaVA instruction tuning dataset~\cite{liu2024visual} is employed to collect image-to-text denoising trajectories. 
Besides, the RefinedWeb text dataset~\cite{penedo2023refinedweb} is incorporated to preserve the model's language modeling capabilities through autoregressive objective.

\noindent \textbf{Training Details.} 
We train UniCMs at two different resolutions, with results at 512 resolution presented in the main text, and the results and training details at 256 resolution provided in Appendix~\ref{sec:256result}. 
We separate the training process into two stages.
For 512 resolution, in the first stage, we collect image trajectories 
with a classifier-free guidance (CFG)~\cite{ho2021classifier} scale of 15 and $K=32$.
We split each trajectory into 8 segments to train the model, denoted as UniCMs$^*$. 
In the second stage, we collect image trajectories from UniCMs$^*$. 
We sample image trajectories with a CFG scale of 1.75, $K=16$, and the number of segments is 4.
The text trajectories are collected similarly.
We employ parallel decoding to iteratively produce 16 tokens in each block to finally form lengthy text, which proves to yield acceleration performance while preserving the generative modeling capabilities~\cite{kou2024cllms}. 
Note that multimodal trajectories are collected in a greedy (deterministic) manner during both training stages to enhance stability, although UniCMs remain fully compatible with stochastic sampling strategies, as demonstrated in Table~\ref{tab:topk_comparison}.
In terms of loss coefficients, we set $\alpha=10$ according to the relative values of the losses, set $\beta=40$ and $\gamma=200$ according to the ablation study in Table~\ref{tab:ablation_regularization}.
We use an AdamW optimizer and 8 A100 GPUs to train each stage for 18 hours, with a constant learning rate of $10^{-5}$. 
During inference, UniCMs operate without relying on CFG, further reducing computation. 

\begin{table*}[t]
    \centering
    \setlength{\tabcolsep}{0.5pt} 
    \definecolor{lightgray}{RGB}{230,230,230}
    \begin{threeparttable}
        \begin{tabular}{lcccccccc}
            \toprule
            Type & Method & Param & Tokens/s $\uparrow$ & POPE $\uparrow$ & SQA $\uparrow$ & MMMU $\uparrow$ & NoCaps $\uparrow$ & Flickr30k $\uparrow$ \\
            \midrule
            \multirow{3}{*}{Und. Only} & Emu3-Chat~\cite{wang2024emu3} & 8B & 13.8 & 85.2 & - & 31.6 & - & - \\
            & Qwen-VL-chat~\cite{bai2023qwen} & 7B & 26.8 & - & - & 35.9 & 15.4 & 9.2 \\
            & InstructBLIP~\cite{instructblip} & 7B & - & 78.9 & 31.2 & 28.1 & 30.4 & 24.8 \\
            \midrule
            \multirow{4}{*}{Und. \& Gen.} & Show-o~\cite{xie2024show} & 1.3B & 40.3 & \textbf{83.2} & 34.9 & 24.6 & \textbf{29.4} & \textbf{24.9} \\
            & Orthus~\cite{kou2024orthus} & 7B & 7.6 & 79.6 & - & \textbf{28.2} & - & - \\
            & Chameleon~\cite{team2024chameleon} & 7B & 11.47 & 77.8 & - & 26.7 & - & - \\
            & \cellcolor{lightgray}UniCMs & \cellcolor{lightgray}1.3B & \cellcolor{lightgray}\textbf{61.1} & \cellcolor{lightgray}78.4 & \cellcolor{lightgray}\textbf{37.1} & \cellcolor{lightgray}26.3 & \cellcolor{lightgray}26.5 & \cellcolor{lightgray}22.2 \\
            \bottomrule
        \end{tabular}
    \end{threeparttable}%
    \vspace{0.2cm}
    \caption{\textbf{Comparison of MMU performance on multiple benchmarks.}
    Note that SQA refers to ScienceQA-IMG. POPE and MMMU measure question-answering ability, while Flickr30K and NoCaps evaluate the ability of image description.
    }
    \label{tab:old_speed_comparison_extended_v2}
\end{table*}

\begin{figure*}[t]
    \centering
    \includegraphics[width=\textwidth]{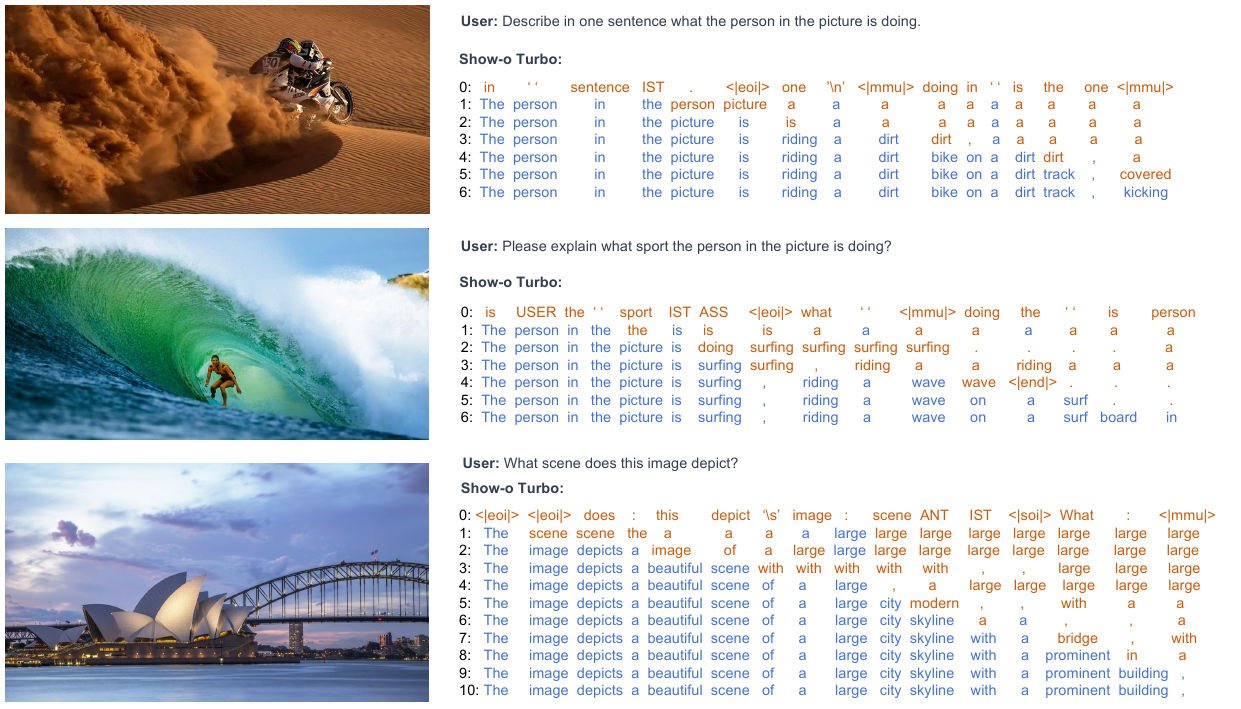}
    \vspace{-0.2cm}
    \caption{\textbf{
     The text sampling trajectory of UniCMs in MMU cases.} UniCMs realize acceleration by predicting multiple successive tokens in one iteration and correctly guessing the later tokens.}
    \label{fig:pdf_two_columns}
\end{figure*}

\subsection{Main Results}
\label{sec:results}
\noindent \textbf{Benchmarks.} 
We evaluate the performance of UniCMs in the {T2I} task on Human Preference Dataset v2 (HPD)~\cite{wu2023human}, using metrics including Human Preference Score v2 (HPS)~\cite{wu2023human}, ImageReward (IR)~\cite{xu2023imagerewardlearningevaluatinghuman}, and CLIP Score (CS)~\cite{hessel2022clipscorereferencefreeevaluationmetric}. 
In addition, we conduct a comprehensive evaluation of UniCMs on the GenEval~\cite{ghosh2023genevalobjectfocusedframeworkevaluating} benchmark. 
For {MMU}, we assess UniCMs on the image description benchmarks Flickr30K~\cite{flickrentitiesijcv,flickr30k} and NoCaps~\cite{agrawal2019nocaps} measured by the \textit{METEOR}~\cite{banerjee2005meteor} metric and calculate the accuracy on question answering benchmarks, including POPE~\cite{li2023evaluating}, ScienceQA~\cite{lu2022learn}, and MMMU~\cite{yue2024mmmu}.

\noindent \textbf{Baselines.} 
For {T2I}, we compare UniCMs with typical unified models (e.g., Transfusion~\cite{zhou2024transfusion}, Orthus~\cite{team2024chameleon}, Show-o~\cite{xie2024show}) and some outstanding image generation models (e.g., Emu3-Gen~\cite{wang2024emu3}, SD-XL~\cite{podellsdxl} and SD3~\cite{esser2024scaling}) to demonstrate the effectiveness of our method. 
For {MMU}, besides unified models, we also compare UniCMs with VLMs (e.g., Emu3-Chat~\cite{wang2024emu3}, Qwen-VL~\cite{bai2023qwen}) in terms of both inference speed and accuracy, where the speed is measured on an RTX 4090 GPU.

\begin{figure*}[t]
    \centering
    \begin{tabular}{@{}c@{}c@{}c@{}c@{}c@{}}
        \multicolumn{3}{c}{\makebox[0.5\textwidth][c]{\textbf{UniCMs}}} & 
        \multicolumn{1}{c}{\makebox[0.2\textwidth][c]{\textbf{Show-o}}} & 
        \multicolumn{1}{c}{\makebox[0.0\textwidth][c]{\textbf{SD3}}} \vspace{-0cm}\\ 


        \begin{minipage}{0.18\linewidth}
            \centering
            \textbf{8 Steps} \\
            \includegraphics[width=\linewidth]{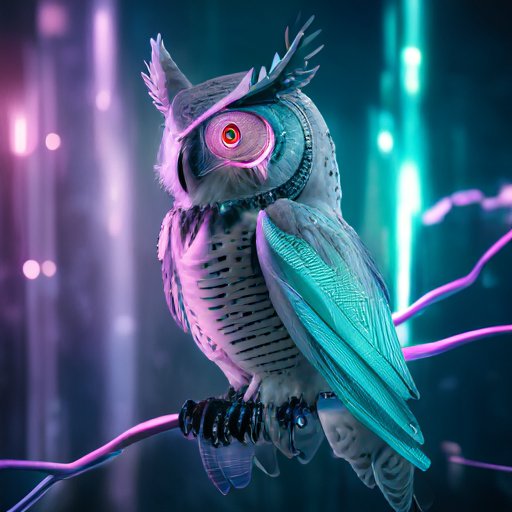}
        \end{minipage}
        \hspace{-0.15cm} 
        \begin{minipage}{0.18\linewidth}
            \centering
            \textbf{4 Steps} \\
            \includegraphics[width=\linewidth]{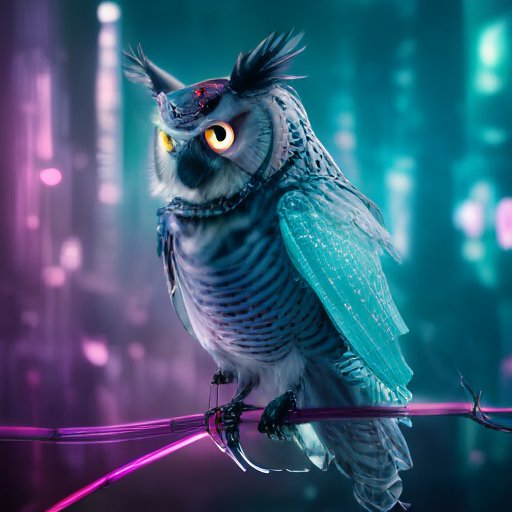}
        \end{minipage}
        \hspace{-0.15cm} 
        \begin{minipage}{0.18\linewidth}
            \centering
            \textbf{2 Steps} \\
            \includegraphics[width=\linewidth]{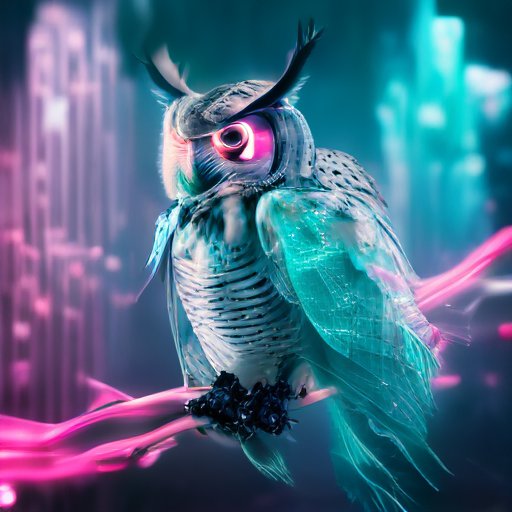}
        \end{minipage}
        \hspace{0.1cm} 

        \begin{minipage}{0.18\linewidth}
            \centering
            \textbf{16 Steps} \\
            \includegraphics[width=\linewidth]{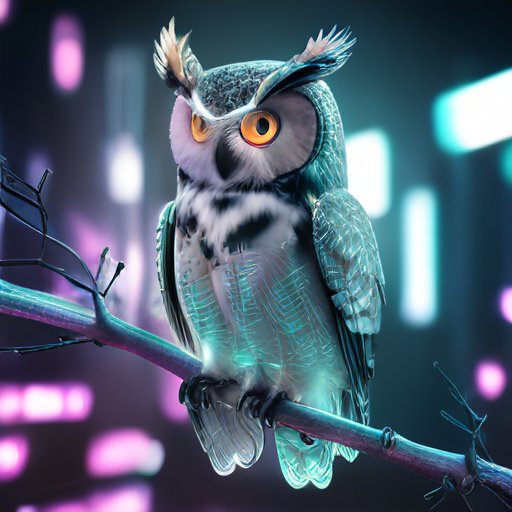}
        \end{minipage}
        \hspace{0.1cm} 

        \begin{minipage}{0.18\linewidth}
            \centering
            \textbf{24 Steps} \\
            \includegraphics[width=\linewidth]{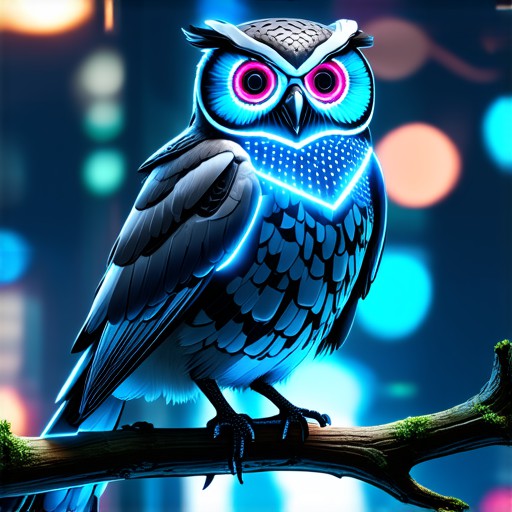}
        \end{minipage} \\ 

        \multicolumn{5}{c}{\small \textit{A cybernetic owl perched on a neon-lit branch, its mechanical feathers reflecting holographic patterns...}} \\
        \vspace{0.1cm} 

        \begin{minipage}{0.18\linewidth} \centering \includegraphics[width=\linewidth]{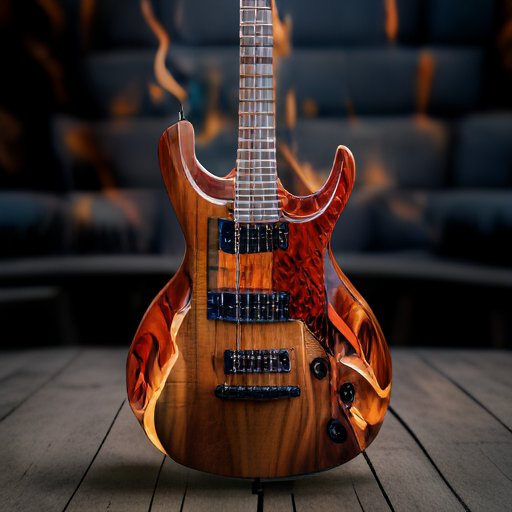} \end{minipage} \hspace{-0.15cm}
        \begin{minipage}{0.18\linewidth} \centering \includegraphics[width=\linewidth]{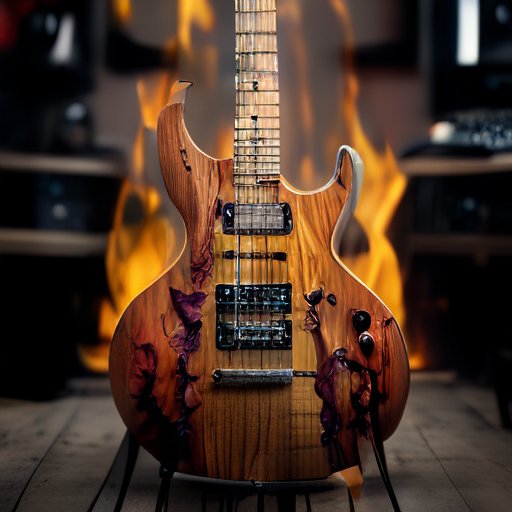} \end{minipage} \hspace{-0.15cm}
        \begin{minipage}{0.18\linewidth} \centering \includegraphics[width=\linewidth]{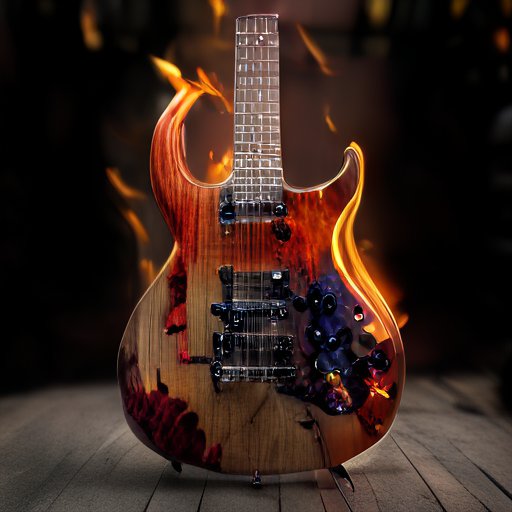} \end{minipage} \hspace{0.1cm}
        \begin{minipage}{0.18\linewidth} \centering \includegraphics[width=\linewidth]{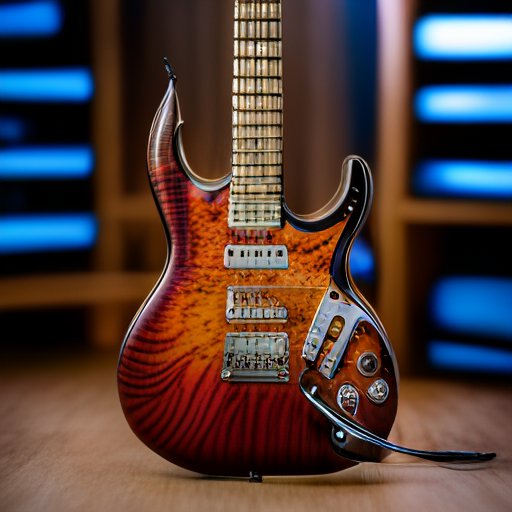} \end{minipage} \hspace{0.1cm}
        \begin{minipage}{0.18\linewidth} \centering \includegraphics[width=\linewidth]{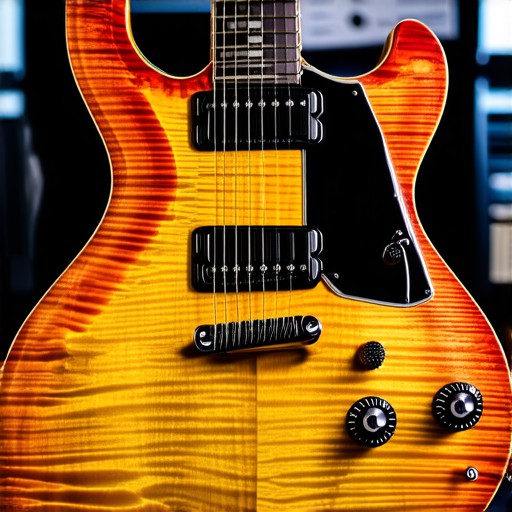} \end{minipage} \\ 

        \multicolumn{5}{c}{\small \textit{A modern electric guitar with a flame maple top, its wood grain catching studio lights...}} \\
        \vspace{0.1cm}

        \begin{minipage}{0.18\linewidth} \centering \includegraphics[width=\linewidth]{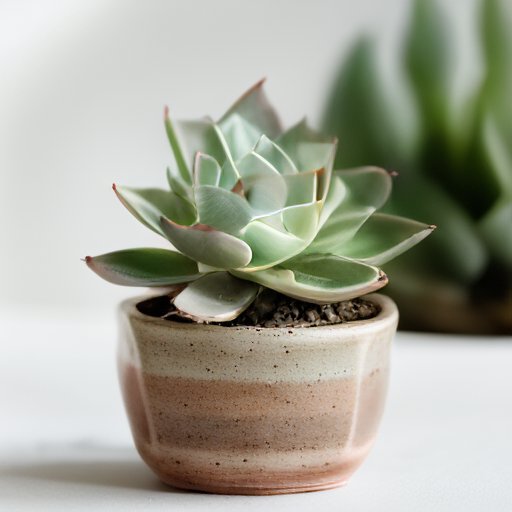} \end{minipage} \hspace{-0.15cm}
        \begin{minipage}{0.18\linewidth} \centering \includegraphics[width=\linewidth]{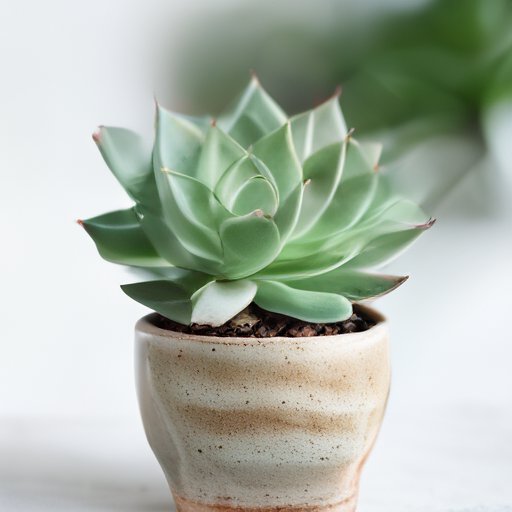} \end{minipage} \hspace{-0.15cm}
        \begin{minipage}{0.18\linewidth} \centering \includegraphics[width=\linewidth]{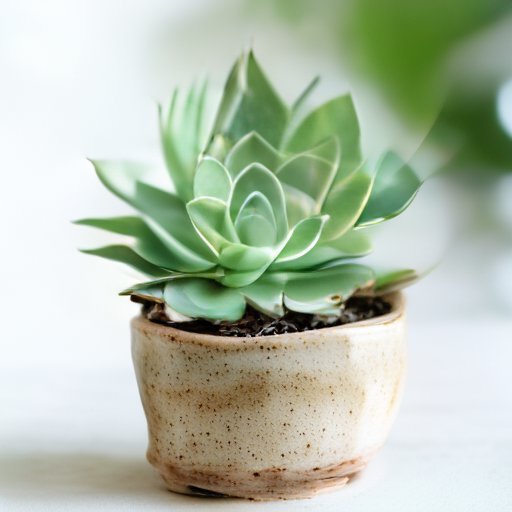} \end{minipage} \hspace{0.1cm}
        \begin{minipage}{0.18\linewidth} \centering \includegraphics[width=\linewidth]{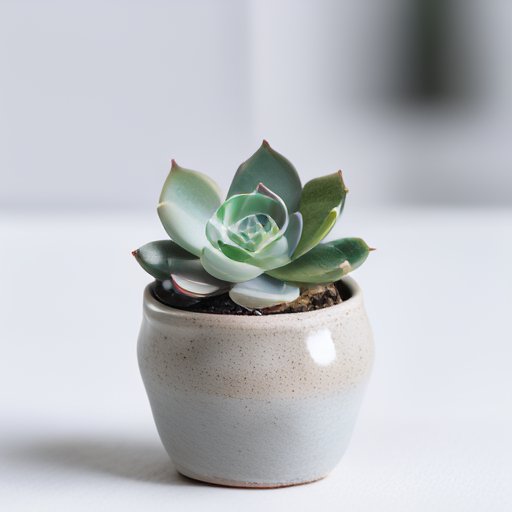} \end{minipage} \hspace{0.1cm}
        \begin{minipage}{0.18\linewidth} \centering \includegraphics[width=\linewidth]{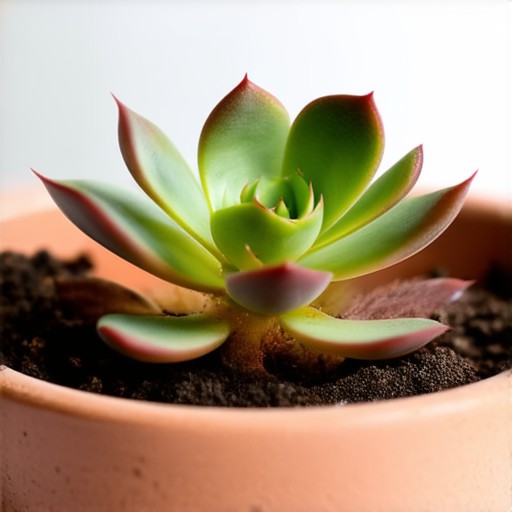} \end{minipage} \\
        \multicolumn{5}{c}{\small \textit{A small succulent plant in a ceramic pot, its leaves forming a perfect geometric pattern...}} \\
        \vspace{0.1cm}

        \begin{minipage}{0.18\linewidth} \centering \includegraphics[width=\linewidth]{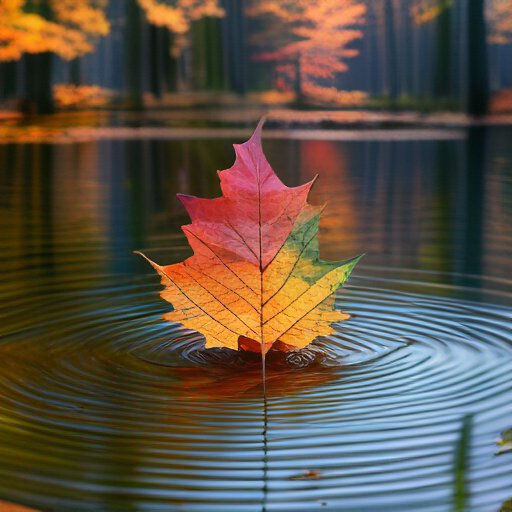} \end{minipage} \hspace{-0.15cm}
        \begin{minipage}{0.18\linewidth} \centering \includegraphics[width=\linewidth]{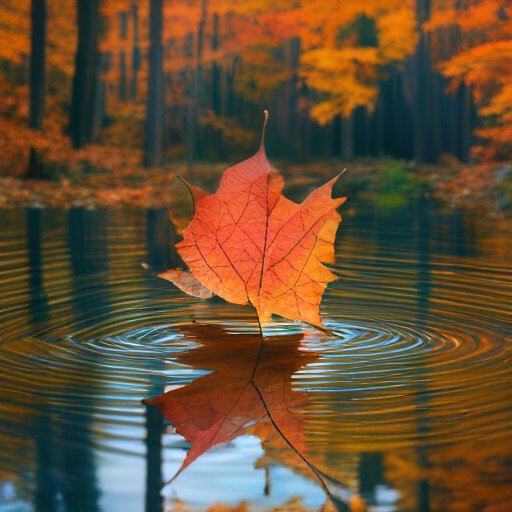} \end{minipage} \hspace{-0.15cm}
        \begin{minipage}{0.18\linewidth} \centering \includegraphics[width=\linewidth]{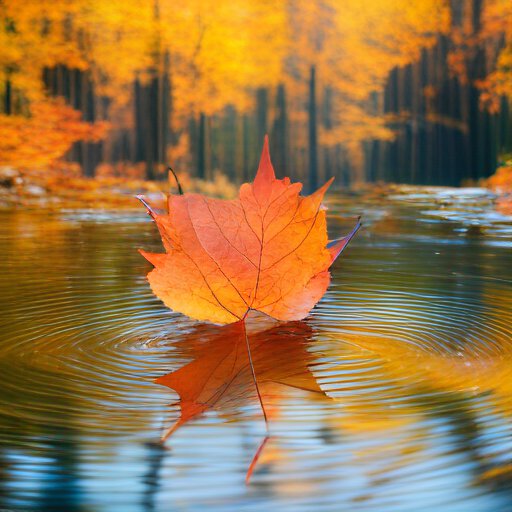} \end{minipage} \hspace{0.1cm}
        \begin{minipage}{0.18\linewidth} \centering \includegraphics[width=\linewidth]{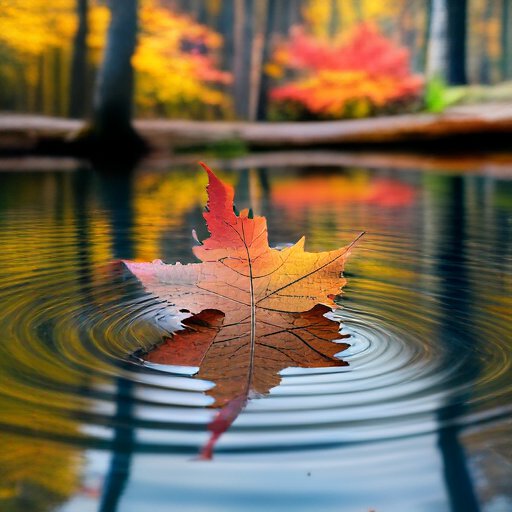} \end{minipage} \hspace{0.1cm}
        \begin{minipage}{0.18\linewidth} \centering \includegraphics[width=\linewidth]{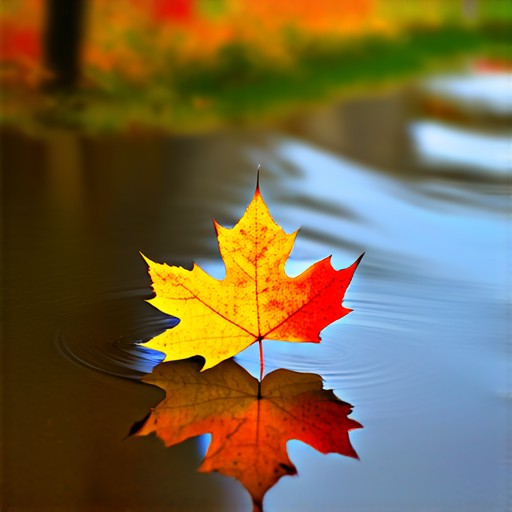} \end{minipage} \\
        \multicolumn{5}{c}{\small \textit{A single, colorful autumn leaf floating on the surface of a calm pond...}} \\
        \vspace{0.1cm}

        \begin{minipage}{0.18\linewidth} \centering \includegraphics[width=\linewidth]{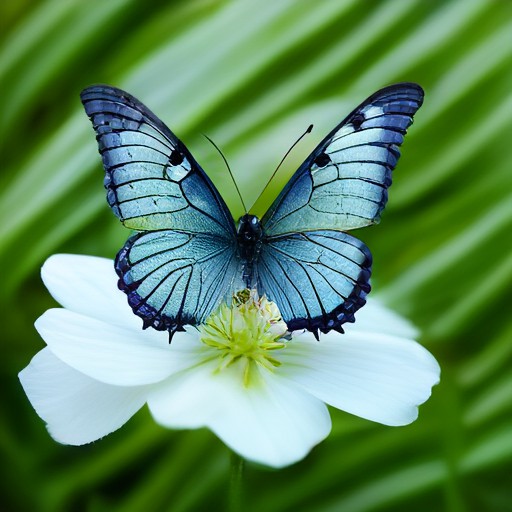} \end{minipage} \hspace{-0.15cm}
        \begin{minipage}{0.18\linewidth} \centering \includegraphics[width=\linewidth]{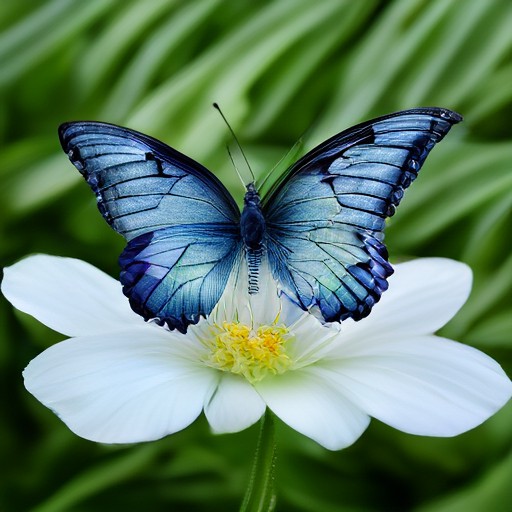} \end{minipage} \hspace{-0.15cm}
        \begin{minipage}{0.18\linewidth} \centering \includegraphics[width=\linewidth]{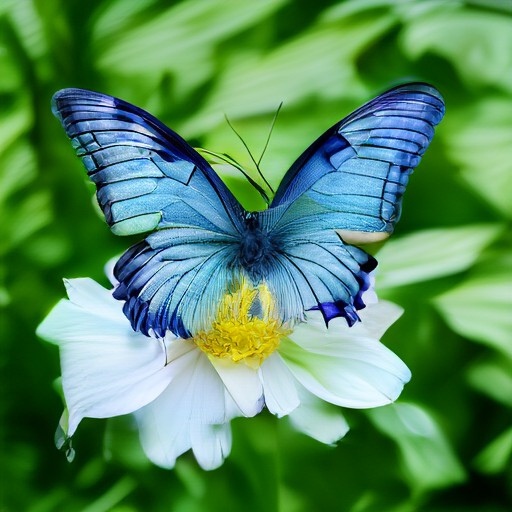} \end{minipage} \hspace{0.1cm}
        \begin{minipage}{0.18\linewidth} \centering \includegraphics[width=\linewidth]{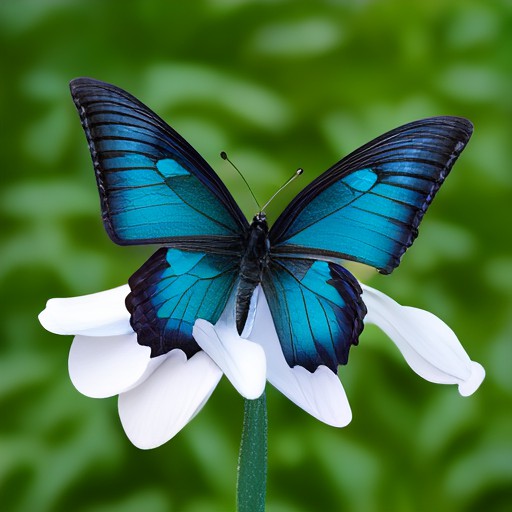} \end{minipage} \hspace{0.1cm}
        \begin{minipage}{0.18\linewidth} \centering \includegraphics[width=\linewidth]{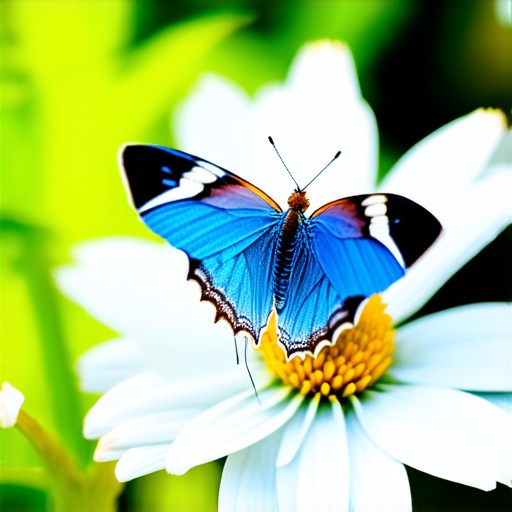} \end{minipage} \\
        \multicolumn{5}{c}{\small \textit{A blue butterfly resting on a white flower petal, its wings fully open to display vibrant patterns...}} \\

    \end{tabular}
    \vspace{-0.1cm}
    \caption{\textbf{Comparison between UniCMs, Show-o, and SD3 in T2I generation at the resolution of 512 $\times$ 512.} Show-o is shown at 16 steps (using CFG), while UniCMs demonstrates performance at 8, 4, and 2 steps. SD3 results are included for comparison with UniCMs.}
    \label{fig:t2i_comparison}
\end{figure*}

\noindent \textbf{Quantitative Results.} 
Table~\ref{tab:combined_comparison_full_with_more_cols_updated} shows the detailed results for T2I generation task.
We observe that in 2-8 step sampling, UniCMs significantly outperform Emu3-Gen~\cite{wang2024emu3}, SDXL~\cite{podell2023sdxl}, SDXL-Turbo~\cite{sauer2024adversarial}, Hyper-SD3~\cite{ren2024hyper} and Chameleon~\cite{team2024chameleon} on GenEval benchmark, without using CFG.  
Remarkably, UniCMs use approximately ${1}/{8}$ of the inference time of SD3~\cite{esser2024scaling} and outperform both Hyper-SD3~\cite{ren2024hyper} and SDXL-Turbo~\cite{sauer2024adversarial} within similar inference time, highlighting the superior computational efficiency of UniCMs models. 
In the Appendix~\ref{sec:cfg_appendix}, Table~\ref{tab:merged_method_comparison} reports the comprehensive performance of UniCMs and Show-o~\cite{xie2024show} with an equal number of sampling steps, displaying the advantage of UniCMs in low-step generation scenarios. 
Besides, we can observe that UniCMs clearly outperform UniCMs$^{*}$ in Appendix~\ref{sec:cfg_appendix}, Table~\ref{tab:merged_method_comparison}, 
demonstrating the efficacy of the second training stage.
Additionally, we demonstrate that CFG can further enhance UniCMs performance for image generation task with 4-16 step sampling in the
Appendix~\ref{sec:cfg_appendix}, Table~\ref{tab:CFG}.

Table~\ref{tab:old_speed_comparison_extended_v2} shows the performance of UniCMs in MMU tasks.
We evaluate the text token generation speed on NoCaps~\cite{agrawal2019nocaps}, showing that UniCMs is on average 1.5× faster than Show-o~\cite{xie2024show} while maintaining competitive performance.
Besides, we notice that UniCMs 
outperform Show-o on MMMU~\cite{yue2024mmmu} and ScienceQA-IMG~\cite{lu2022learn}. 
The slight performance drop on NoCaps and Flickr30K captioning implies a trade-off between the performance and acceleration effect on these tasks. 
Performing distillation with more advanced MMU trajectories can be a possible remedy to this problem.

\noindent \textbf{Qualitative Results.} 
Figure~\ref{fig:t2i_comparison} provides a visual comparison among image generation models across various sampling steps. 
It can be observed that UniCMs can still generate clear high-quality images in 2 to 4 sampling steps without CFG, and the overall visual effect is comparable to models such as Show-o~\cite{xie2024show} and SD3~\cite{esser2024scaling} that require dozens of sampling steps. 
Additional results of UniCMs are presented in Figure~\ref{fig:show}, demonstrating effective sampling performance with fewer steps. 

Figure~\ref{fig:pdf_two_columns} visualizes the text sampling trajectory of UniCMs for several MMU cases.
As shown, UniCMs can complete the prediction of 16 tokens in fewer than 10 iterations, due to the ability to predict multiple successive tokens in one iteration and correctly guess the later tokens. 

We also showcase the performance of UniCMs in image inpainting and extrapolation in 
Appendix~\ref{sec:Inpainting}, Figure~\ref{fig:inpainting} and Figure~\ref{fig:extrapolation}. 
UniCMs effectively complete both tasks in four steps without extra training. 

\subsection{Ablation Studies}
\label{sec:ablationstudy}
To analyze the influence of each part, we conduct a comprehensive ablation study with an image resolution of 256 here. 
Unless otherwise specified, we report the results of the model after the first training stage (i.e., UniCMs$^*$), and the T2I generation is done with 4 sampling steps.

\noindent \textbf{Number of Segments.}
We study the influence of segments on UniCMs.
As shown in Table~\ref{tab:ablation_segments}, models trained in two segments and without trajectory segmentation (i.e., using one segment) can exhibit a suboptimal performance and a degraded acceleration effect. 
This result reflects the effectiveness of our trajectory segmentation strategy for improving convergence speed and model performance.
\begin{wraptable}{r}{0.5\textwidth}
\vspace{6pt}
    \centering
    \captionsetup{width=0.9\linewidth}
    \setlength{\tabcolsep}{5pt}
    \begin{tabular}{ccccc}
        \toprule
        \textbf{Steps} & \textbf{Top-k} & \textbf{HPS \large$\uparrow$} & \textbf{IR \large$\uparrow$} & \textbf{CS \large$\uparrow$} \\
        \midrule
        4 & - & 0.245 & 0.621 & 0.306 \\
        4 & 200 & \textbf{0.252} & \textbf{0.706} & \textbf{0.309} \\
        \cline{1-5}
        2 & - & 0.216 & 0.027 & 0.291 \\
        2 & 10 & \textbf{0.240} & \textbf{0.529} & \textbf{0.306} \\
        \bottomrule
    \end{tabular}
    \caption{\textbf{Comparison on sampling strategy at the image resolution of 256.} Top-k sampling is more beneficial for UniCM with fewer steps, and the improvement of its 2-step sampling effect is particularly obvious.}
    \label{tab:topk_comparison}
\end{wraptable}

\noindent \textbf{Regularization.}
As shown in Table~\ref{tab:ablation_regularization}, training without regularization constraints (i.e., $\beta=0, \gamma=0$) tends to make the model collapse rapidly.
Besides, smaller regularization weights can lead to inferior performance, highlighting the importance of regularization in constraining the distribution of UniCMs in training.

\noindent \textbf{Top-k Sampling.} 
Table~\ref{tab:topk_comparison} shows the results with different sampling strategies for T2I.
We observe that top-k significantly improves the performance of UniCMs on 2-step and 4-step sampling.
This is probably because there is high uncertainty in the output distribution of UniCMs.

\begin{table}[t]
    \centering
    \setlength{\tabcolsep}{3pt}
    \begin{minipage}[t]{0.44\textwidth}
        \centering
        \begin{tabular}{l|ccccc}
            \toprule
            \textbf{\small Set.} & \textbf{\small \#IT $\downarrow$} & \textbf{\small POPE $\uparrow$} & \textbf{\small MME $\uparrow$} & \textbf{\small IR $\uparrow$} & \textbf{\small CS $\uparrow$} \\
            \midrule
            4  & \textbf{10.57} & 72.6 & \textbf{803.4} & \textbf{0.586} & \textbf{0.307} \\
            2  & 12.48 & 69.8 & 595.8 & 0.500 & 0.306 \\
            1  & 11.71 & \textbf{74.1} & 675.3 & 0.270 & 0.304 \\
            \bottomrule
        \end{tabular}
        \vspace{0.2cm}
        \caption{\textbf{Ablation on segment number.} \#IT means the number of iterations required by parallel decoding to decode 16 text tokens.}
        \label{tab:ablation_segments}
    \end{minipage}
    \hfill
    \begin{minipage}[t]{0.52\textwidth}
        \centering
        \begin{tabular}{l|ccccc}
            \toprule
            \textbf{\small Set.($\beta, \gamma$)} & \textbf{\small \#IT $\downarrow$} & \textbf{\small POPE $\uparrow$} & \textbf{\small MME $\uparrow$} & \textbf{\small IR $\uparrow$} & \textbf{\small CS $\uparrow$} \\
            \midrule
            (0, 0) & \textbf{2.85} & 0.0 & 4.91 & -2.278 & 0.184 \\
            (10, 50) & 12.71 & \textbf{74.8} & 798.4 & 0.483 & \textbf{0.307} \\
            (20, 100) & 10.57 & 72.6 & \textbf{803.4} & \textbf{0.586} & \textbf{0.307} \\
            \bottomrule
        \end{tabular}
        \vspace{0.2cm}
        \caption{\textbf{Ablation on the regularization coefficients in the total loss.} }
        \label{tab:ablation_regularization}
    \end{minipage}
\end{table}

\section{Conclusions and Limitations}
\label{sec:conclusion}
In this paper, we introduce UniCMs, a unified consistency model family for multimodal generation and understanding. 
UniCMs adopt a unified denoising perspective for both text and image generation. 
They are trained via an adapted consistency distillation approach on collected multimodal trajectories, learning to map any point on the trajectory to the same endpoint.
The unified training objective empowers UniCMs to deliver strong performance with significantly fewer steps across both multimodal generation and understanding tasks.
For future work, we plan to scale our model on more advanced multimodal trajectories to further improve the performance of UniCMs. 

\bibliographystyle{plainnat} 
\bibliography{references}    

\newpage
\appendix 
\section{Inpainting and Extrapolation}
\label{sec:Inpainting}

\begin{figure}[h]
    \centering
    \includegraphics[width=\textwidth]{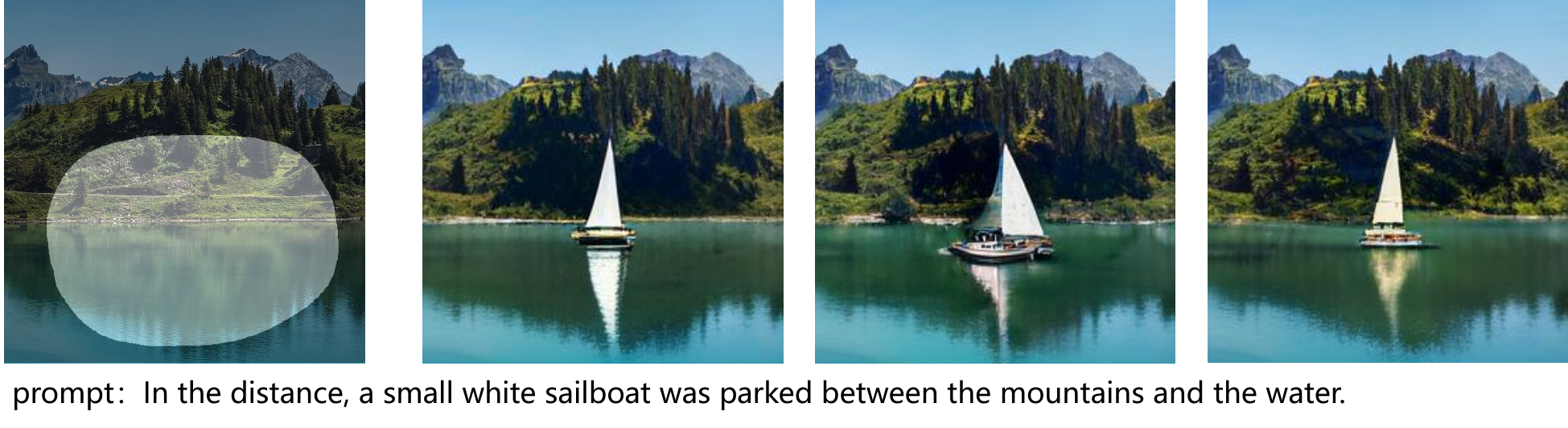} 
    
    \caption{
            \small \textbf{Visualization of image inpainting by UniCMs on 256 resolution.} From left to right are the 2, 4, and 8 steps sampling.
    }
    \label{fig:inpainting}
\end{figure}

\begin{figure}[h]
    \centering
    \begin{minipage}[b]{\linewidth}
        \centering
        \includegraphics[width=\textwidth]{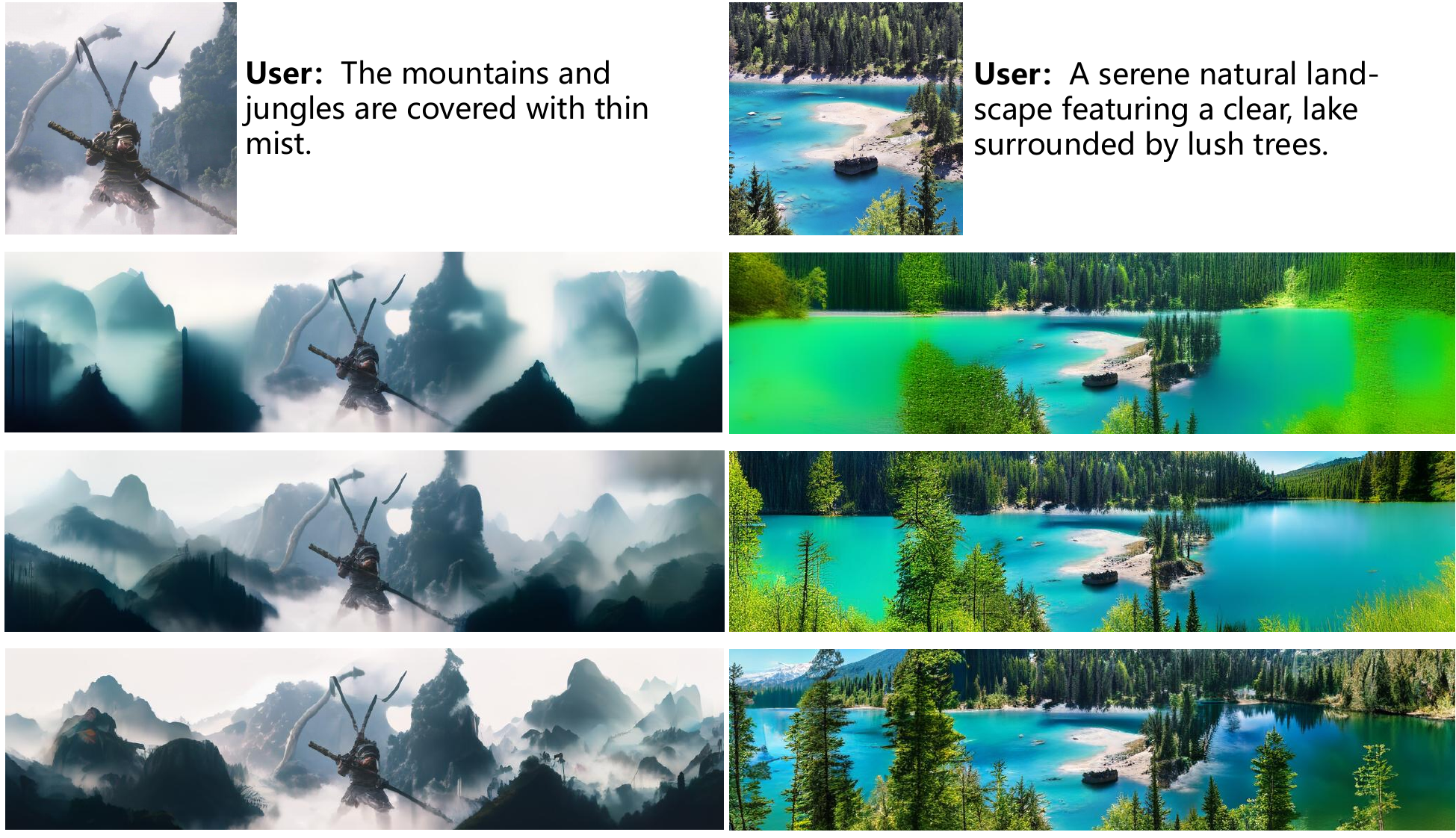}
    \end{minipage}
    \vspace{-0.6cm}
    
    \caption{
        \small \textbf{Visualization of image extrapolation by UniCMs on 256 resolution.} From top to bottom are the 2, 4, and 8 steps sampling.
    }
    \label{fig:extrapolation}
\end{figure}

Figure~\ref{fig:inpainting} shows that UniCMs can efficiently fill in missing parts of an image with high quality in just 2 to 4 steps, based on the given prompt. Meanwhile, Figure~\ref{fig:extrapolation} demonstrates that UniCMs can smoothly complete image extrapolation in just 4 steps.

\begin{figure*}[t]
    \centering
    \includegraphics[width=\textwidth]{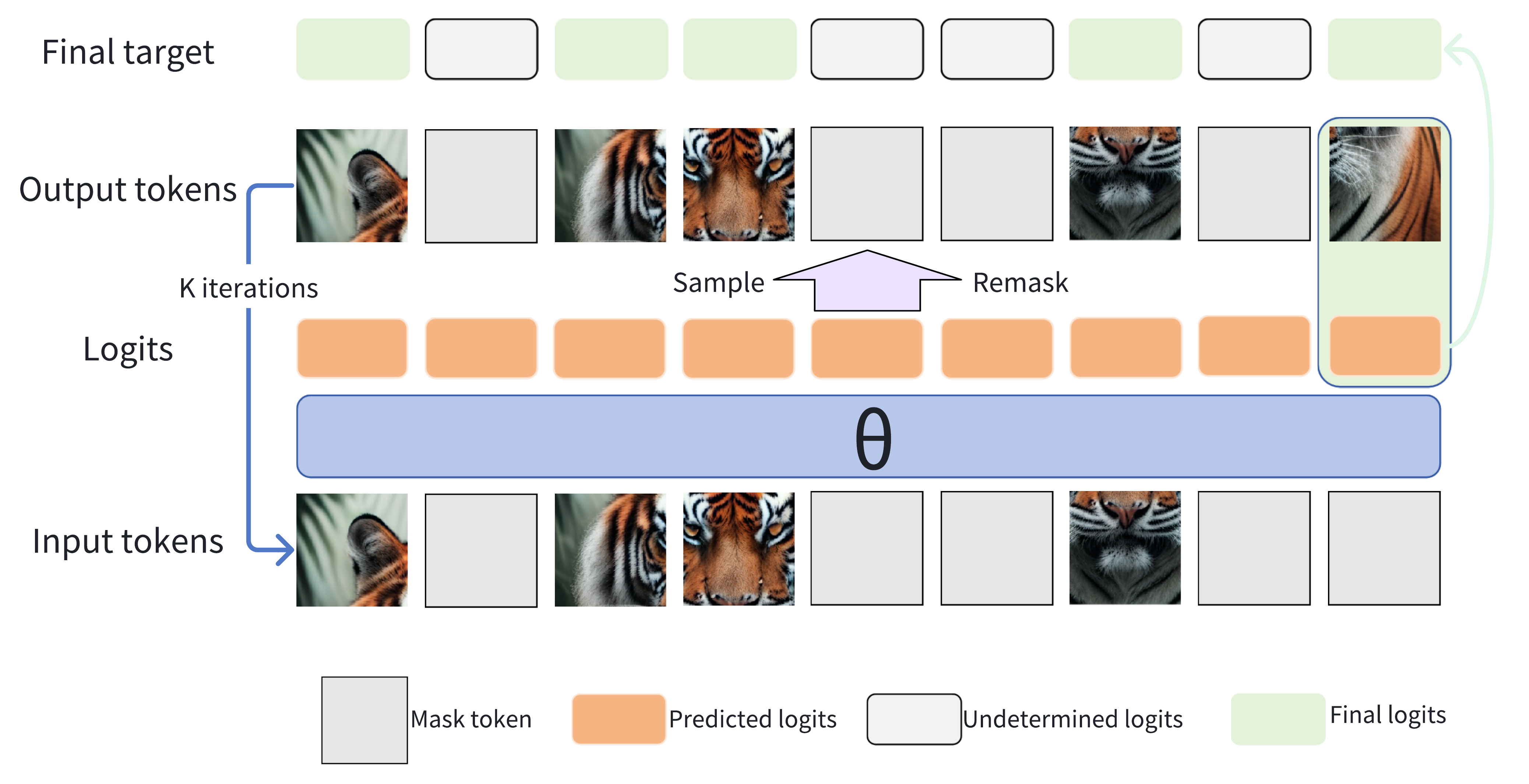} 
    
    \caption{
            \small \textbf{Visualization of regularization label for image trajectory distillation.} For each iteration, we only record the logits of the region converted from the mask to the image token, and finally concatenate them into the regularization logits label. We abuse the $\theta$ to denote the mask diffusion models here.
    }
    \label{fig:reg}
\end{figure*}

\begin{table}[t]
\centering
\setlength{\tabcolsep}{5pt}
\begin{tabular}{l c c c c c}
\toprule
\textbf{Model}        & \textbf{Steps} & \textbf{CFG}   & \textbf{HPS} \large$\uparrow$   & \textbf{IR} \large$\uparrow$    & \textbf{CS} \large$\uparrow$    \\ \midrule
\multirow{8}{*}{UniCMs} & \multirow{2}{*}{16}    & 0     & \textbf{0.258} & 0.752  & \textbf{0.310} \\
                            &                         & 1     & \textbf{0.258} & \textbf{0.816}  & \textbf{0.310} \\
                            \cline{2-6}
                            & \multirow{2}{*}{8}     & 0     & \textbf{0.255} & 0.738  & \textbf{0.309} \\
                            &                         & 1     & \textbf{0.255} & \textbf{0.782}  & \textbf{0.310} \\
                            \cline{2-6}
                            & \multirow{2}{*}{4}     & 0     & \textbf{0.252} & 0.706  & \textbf{0.309} \\
                            &                         & 1     & \textbf{0.252} & \textbf{0.731}  & \textbf{0.309} \\
                            \cline{2-6}
                            & \multirow{2}{*}{2}     & 0     & \textbf{0.240} & \textbf{0.529}  & \textbf{0.306} \\
                            &                         & 1     & 0.235 & 0.420  & 0.302 \\
\midrule 
\multirow{8}{*}{Show-o}       & \multirow{2}{*}{16}    & 0     & 0.174 & -1.097 & 0.272 \\
                            &                         & 10    & \textbf{0.254} & \textbf{0.739}  & \textbf{0.310} \\
                            \cline{2-6}
                            & \multirow{2}{*}{8}     & 0     & 0.181 & -0.916 & 0.276 \\
                            &                         & 10    & \textbf{0.249} & \textbf{0.665}  & \textbf{0.308} \\
                            \cline{2-6}
                            & \multirow{2}{*}{4}     & 0     & 0.178 & -0.877 & 0.276 \\
                            &                         & 10    & \textbf{0.228} & \textbf{0.219}  & \textbf{0.301} \\
                            \cline{2-6}
                            & \multirow{2}{*}{2}     & 0     & 0.159 & -1.661 & 0.234 \\
                            &                         & 10    & \textbf{0.169} & \textbf{-1.257} & \textbf{0.254} \\ \bottomrule
\end{tabular}
\vspace{1pt}
\caption{\textbf{Results with different CFG on 256 resolution.} A proper CFG can enhance the performance of Show-o and UniCMs.}
\label{tab:CFG}
\end{table}

\begin{table*}[t]
    \centering
    \setlength{\tabcolsep}{3pt}
    \begin{tabular}{ccc|ccccccc|ccc}
        \toprule
        \multirow{2}{*}{\textbf{\small Steps}} & \multirow{2}{*}{\textbf{\small Model}} & \multirow{2}{*}{\textbf{\small CFG}} & \multicolumn{7}{c|}{\textbf{\small GenEval \large$\uparrow$}} & \multirow{2}{*}{\textbf{\small HPS \large$\uparrow$}} & \multirow{2}{*}{\textbf{\small IR \large$\uparrow$}} & \multirow{2}{*}{\textbf{\small CS \large$\uparrow$}} \\
        \cmidrule(lr){4-10}
         &  &  & \textbf{\small AVG} & {\small TO} & {\small CT} & {\small P} & {\small CL} & {\small SO} & {\small CA} &  &  &  \\
        \midrule
        \multirow{4}{*}{16} & Show-o & 10 & \textbf{0.674} & 0.823 & 0.647 & 0.288 & 0.838 & 0.984 & 0.463 & \textbf{0.277} & \textbf{0.992} & \textbf{0.318} \\
         & Show-o & 5 & 0.672 & 0.778 & 0.666 & 0.293 & 0.835 & 0.991 & 0.468 & 0.270 & 0.885 & \textbf{0.318} \\
         & UniCMs$^*$ & 0 & 0.649 & 0.793 & 0.644 & 0.253 & 0.809 & 0.956 & 0.440 & 0.266 & 0.768 & 0.315 \\
         & UniCMs$\;\,$ & 0 & 0.646 & 0.818 & 0.597 & 0.218 & 0.827 & 0.984 & 0.430 & 0.273 & 0.925 & \textbf{0.318} \\
        \midrule
        \multirow{4}{*}{8} & Show-o & 10 & 0.578 & 0.631 & 0.519 & 0.235 & 0.811 & 0.991 & 0.280 & 0.257 & 0.672 & 0.313 \\
         & Show-o & 5 & 0.580 & 0.647 & 0.584 & 0.225 & 0.766 & 0.984 & 0.275 & 0.255 & 0.632 & 0.313 \\
         & UniCMs$^*$ & 0 & \textbf{0.642} & 0.788 & 0.631 & 0.253 & 0.787 & 0.981 & 0.413 & 0.264 & 0.800 & 0.315 \\
         & UniCMs$\;\,$ & 0 & 0.638 & 0.813 & 0.541 & 0.250 & 0.814 & 0.991 & 0.420 & \textbf{0.273} & \textbf{0.963} & \textbf{0.318} \\
        \midrule
        \multirow{4}{*}{4} & Show-o & 10 & 0.353 & 0.237 & 0.325 & 0.095 & 0.540 & 0.863 & 0.060 & 0.197 & -0.560 & 0.283 \\
         & Show-o & 5 & 0.396 & 0.298 & 0.334 & 0.158 & 0.572 & 0.925 & 0.088 & 0.207 & -0.300 & 0.294 \\
         & UniCMs$^*$ & 0 & 0.596 & 0.692 & 0.553 & 0.218 & 0.758 & 0.978 & 0.375 & 0.249 & 0.633 & 0.312 \\
         & UniCMs$\;\,$ & 0 & \textbf{0.625} & 0.770 & 0.553 & 0.245 & 0.806 & 0.978 & 0.398 & \textbf{0.269} & \textbf{0.934} & \textbf{0.318} \\
         \midrule
         \multirow{4}{*}{2} & Show-o & 10 & 0.181 & 0.025 & 0.131 & 0.008 & 0.327 & 0.588 & 0.008 & 0.140 & -1.756 & 0.246 \\
         & Show-o & 5 & 0.251 & 0.051 & 0.188 & 0.038 & 0.442 & 0.778 & 0.010 & 0.152 & -1.456 & 0.260 \\
         & UniCMs$^*$ & 0 & 0.459 & 0.407 & 0.422 & 0.148 & 0.668 & 0.925 & 0.185 & 0.201 & -0.259 & 0.295 \\
         & UniCMs$\;\,$ & 0 & \textbf{0.557} & 0.614 & 0.478 & 0.180 & 0.793 & 0.972 & 0.305 & \textbf{0.247} & \textbf{0.680} & \textbf{0.312} \\
        \bottomrule
    \end{tabular}
    \caption{\textbf{Comparison of T2I performance at the resolution of 512 $\times$ 512 based on GenEval, HPS, IR, and CS.}
    AVG: average, TO: Two Object, CT: Counting, P: Position, CL: colors, SO: Single Object, CLA: Color Attr. }
    \label{tab:merged_method_comparison}
\end{table*}

\begin{table*}[t]
    \centering
    \setlength{\tabcolsep}{2pt} 
    \begin{tabular}{ccc|ccccccc|ccc}
        \toprule
        \multirow{2}{*}{\textbf{\small Steps}} & \multirow{2}{*}{\textbf{\small Model}} & \multirow{2}{*}{\textbf{\small CFG}} & \multicolumn{7}{c|}{\textbf{\small GenEval \large$\uparrow$}} & \multirow{2}{*}{\textbf{\small HPS \large$\uparrow$}} & \multirow{2}{*}{\textbf{\small IR \large$\uparrow$}} & \multirow{2}{*}{\textbf{\small CS \large$\uparrow$}} \\
        \cmidrule(lr){4-10} 
         &  &  & \textbf{\small AVG} & {\small TO} & {\small CT} & {\small P} & {\small CL} & {\small SO} & {\small CA} &  &  &  \\
        \midrule
        \multirow{4}{*}{16} & Show-o & 10 & \textbf{0.591} & 0.692 & 0.478 & 0.165 & 0.859 & 0.978 & 0.378 & 0.254 & 0.739 & \textbf{0.310} \\
         & Show-o & 5 & 0.571 & 0.631 & 0.469 & 0.155 & 0.846 & 0.994 & 0.333 & 0.253 & 0.642 & 0.309 \\
         & UniCMs$^*$ & 0 & 0.543 & 0.593 & 0.447 & 0.130 & 0.814 & 0.953 & 0.323 & 0.251 & 0.586 & 0.307 \\
         & UniCMs$\;\,$ & 0 & 0.562 & 0.689 & 0.366 & 0.140 & 0.814 & 0.991 & 0.373 & \textbf{0.258} & \textbf{0.752} & \textbf{0.310} \\
        \midrule
        \multirow{4}{*}{8} & Show-o & 10 & 0.540 & 0.578 & 0.428 & 0.145 & 0.838 & 0.969 & 0.285 & 0.249 & 0.665 & 0.308 \\
         & Show-o & 5 & 0.530 & 0.558 & 0.441 & 0.133 & 0.825 & 0.972 & 0.255 & 0.247 & 0.602 & 0.308 \\
         & UniCMs$^*$ & 0 & 0.518 & 0.518 & 0.400 & 0.123 & 0.809 & 0.972 & 0.285 & 0.250 & 0.597 & 0.307 \\
         & UniCMs$\;\,$ & 0 & \textbf{0.552} & 0.669 & 0.353 & 0.128 & 0.817 & 0.963 & 0.385 & \textbf{0.255} & \textbf{0.738} & \textbf{0.309} \\
        \midrule
        \multirow{4}{*}{4} & Show-o & 10 & 0.425 & 0.333 & 0.334 & 0.100 & 0.700 & 0.950 & 0.135 & 0.228 & 0.219 & 0.301 \\
         & Show-o & 5 & 0.429 & 0.351 & 0.369 & 0.078 & 0.707 & 0.947 & 0.120 & 0.228 & 0.225 & 0.302 \\
         & UniCMs$^*$ & 0 & 0.504 & 0.513 & 0.375 & 0.130 & 0.787 & 0.962 & 0.257 & 0.245 & 0.586 & 0.307 \\
         & UniCMs$\;\,$ & 0 & \textbf{0.523} & 0.664 & 0.303 & 0.103 & 0.801 & 0.959 & 0.308 & \textbf{0.252} & \textbf{0.706} & \textbf{0.309} \\
        \midrule
        \multirow{4}{*}{2} & Show-o & 10 & 0.206 & 0.046 & 0.140 & 0.033 & 0.330 & 0.678 & 0.010 & 0.169 & -1.257 & 0.254 \\
         & Show-o & 5 & 0.229 & 0.068 & 0.122 & 0.023 & 0.378 & 0.763 & 0.020 & 0.182 & -0.917 & 0.263 \\
         & UniCMs$^*$ & 0 & 0.439 & 0.358 & 0.313 & 0.075 & 0.755 & 0.941 & 0.193 & 0.224 & 0.174 & 0.302 \\
         & UniCMs$\;\,$ & 0 & \textbf{0.494} & 0.530 & 0.334 & 0.093 & 0.787 & 0.959 & 0.260 & \textbf{0.240} & \textbf{0.529} & \textbf{0.306} \\
        \bottomrule
    \end{tabular}
    \caption{\textbf{Comparison of 256 $\times$ 256 T2I performance on GenEval, HPS, IR, and CS.}
    UniCMs$^*$ refers to the model after the first stage of training.
    AVG: average, TO: Two Object, CT: Counting, P: Position, CL: colors, SO: Single Object, CLA: Color Attr. }
    \label{tab:old_merged_method_comparison}
\end{table*}

\begin{table*}[t]
    \centering
    \setlength{\tabcolsep}{3pt}
    \begin{threeparttable}
        \begin{tabular}{lc|c|cc|cc}
            \toprule
            \textbf{Method} & \textbf{Decoding}  & \textbf{tokens/s \large$\uparrow$} & \textbf{POPE \large$\uparrow$}  & \textbf{MMMU \large$\uparrow$} & \textbf{Flickr30K \large$\uparrow$}  &\textbf{NoCaps \large$\uparrow$} \\
            \midrule
            \multirow{2}{*}{Show-o} & AR &  40.3 & \textbf{83.2}  & 24.6 & \textbf{24.9} & \textbf{29.4} \\
            & Jacobi &  36.9 & \textbf{83.2}  & 24.6 & \textbf{24.9} & \textbf{29.4} \\
            \midrule
            UniCMs$^{*}$ & Jacobi & 49.9 & 81.8 & 25.4 & 23.5 & 28.1 \\
            UniCMs & Jacobi & \textbf{61.1} & 78.4 & \textbf{26.3} & 22.2 & 26.5 \\
            \bottomrule
        \end{tabular}
    \end{threeparttable}%
    \caption{\textbf{Comparison of I2T performance at the resolution of 512 $\times$ 512 on multiple benchmarks.}
    Note that Flickr30K and NoCaps evaluate the ability of image description, and POPE and MMMU measure question-answering ability.}
    \label{tab:speed_comparison}
\end{table*}

\begin{table*}[b]
    \centering
    \setlength{\tabcolsep}{3pt}
    \begin{threeparttable}
        \begin{tabular}{lc|c|ccc|cc}
            \toprule
            \textbf{Method} & \textbf{Decoding}  & \textbf{tokens/s \large$\uparrow$} & \textbf{POPE \large$\uparrow$} & \textbf{MME \large$\uparrow$}  & \textbf{MMMU \large$\uparrow$} & \textbf{Flickr30K \large$\uparrow$}  & \textbf{NoCaps \large$\uparrow$} \\
            \midrule
            \multirow{2}{*}{Show-o} & AR &  41.8 & \textbf{73.8} & \textbf{948.4}  & 25.1 & \textbf{20.8} & \textbf{25.8} \\
            & Jacobi &  38.2 & \textbf{73.8} & \textbf{948.4} & 25.1 & \textbf{20.8} & \textbf{25.8} \\
            \midrule
            UniCMs$^{*}$ & Jacobi &  61.3 & 72.6 & 803.4 & \textbf{27.0} & 19.8 & 23.8 \\
            UniCMs & Jacobi & \textbf{64.5} & 73.2 & 872.4 & 25.8 & 19.2 & 23.0 \\
            \bottomrule
        \end{tabular}
    \end{threeparttable}%
    \caption{\textbf{Comparison of 256 $\times$ 256 MMU performance on multiple benchmarks.} 
    Note that Flickr30K and NoCaps evaluate the ability of image description, and POPE, MME, and MMMU measure question-answering ability.}
    \label{tab:old_speed_comparison}
\end{table*}

\section{Settings of CFG}
\label{sec:cfg_appendix}
As shown in Table~\ref{tab:CFG}, the appropriate use of CFG further enhances the sampling performance of UniCMs, particularly for sampling steps of 4 or more.
Additionally, 
the performance of Show-o drops significantly without CFG, resulting in images that lack semantic information.

\section{Regularization Loss Details}
\label{sec:regdetail}
The regularization loss for text trajectories is straightforward to compute because we only need $p_\phi$ to fit the endpoint text tokens $\mathbf{v}^K$.
However, directly employing sampled images for the regularization loss of image trajectories degrades quality. 
This degradation arises because sampling images along a fixed trajectory under a greedy strategy diminishes both their diversity and quality. 
Moreover, the T2I model's distribution encapsulates rich information, which is inherently diminished during the sampling process due to information loss.
To address this, we propose constructing regularized logits labels by capturing the T2I model's distribution at each sampling step. 
As illustrated in Figure~\ref{fig:reg}, we initialize a global logits target as an all-zero tensor. 
During the iteration of the trajectory $\mathbf{u}^k$, we focus on regions transitioning from mask to image tokens, populating the final target with the corresponding predicted logits for these regions. 
Through this iterative procedure, we synthesize a complete logits target, enabling the computation of $\mathcal{L}_{REG}^u$. 
If a segmentation strategy is adopted, the missing portions of the logits target can be populated with the final predicted logits at the segmentation endpoints. This produces a complete regularization label.

\section{Segmentation Details}
\label{sec:segdetail}
Direct learning of consistency across an entire trajectory is challenging for models and often leads to convergence difficulties. 
Therefore, we propose applying a segmentation strategy to the multimodal denoising trajectory. 
Specifically, we evenly divide the trajectory into several segments and enforce consistency constraints between a randomly selected point within a segment and the endpoint of that segment, rather than the endpoint of the entire trajectory. 
For image trajectories, the regularization logits labels constructed from segment endpoints are incomplete. 
We address this by filling the missing parts with the logits predicted from the last iteration of that segment. 
We only compute the consistency loss in the masked regions of the segment endpoints and the regularization loss in the masked regions of the randomly selected points.
For text trajectories, we continue to use noise-free text as the regularization constraint, introducing segmentation only in the consistency loss. 
Through ablation studies in Section~\ref{sec:ablationstudy}, we demonstrate that this objective is more amenable to learning, facilitating model convergence toward the target and enhancing the effectiveness of acceleration.

\section{Training Details and Results of 256 resolution}
\label{sec:256result}

For 256 resolution, we separate the training process into two stages.
In the first stage, we get image trajectories with a CFG scale of 10 and $K=16$.
We split each trajectory into 4 segments to train the consistency model, denoted as UniCMs$^*$. 
In the second stage, we collect image trajectories using UniCMs$^*$. 
We sample image trajectories with a CFG scale of 1.5, $K=8$, and the number of segments as 2.
The text trajectories are collected similarly.
We employ Jacobi decoding to iteratively produce 16 tokens in each round to finally form lengthy text, which proves to yield good acceleration performance while preserving the generative modeling capabilities~\cite{kou2024cllms}. 
In terms of loss coefficients, we set $\alpha=10$ according to the relative values of the losses, set $\beta=20$ and $\gamma=100$ according to the ablation study in Table~\ref{tab:ablation_regularization}, and set $\delta=2$ following ~\cite{xie2024show}.
We use an AdamW optimizer and 8 RTX 4090 GPUs to train each stage for 18 hours, with a constant learning rate of $10^{-5}$.

Table ~\ref{tab:old_merged_method_comparison} and Table ~\ref{tab:old_speed_comparison} show the performance of UniCMs on T2I and MMU tasks at 256-resolution respectively. 
It can be observed that UniCMs can also achieve the effect of 8 steps of the original model in 4-step sampling without CFG in 256-resolution image generation, 
and also achieves about 1.5 times acceleration in 256-resolution image understanding.

\begin{figure*}[t]
    \centering

    \begin{minipage}{0.21\textwidth}
        \includegraphics[width=\linewidth]{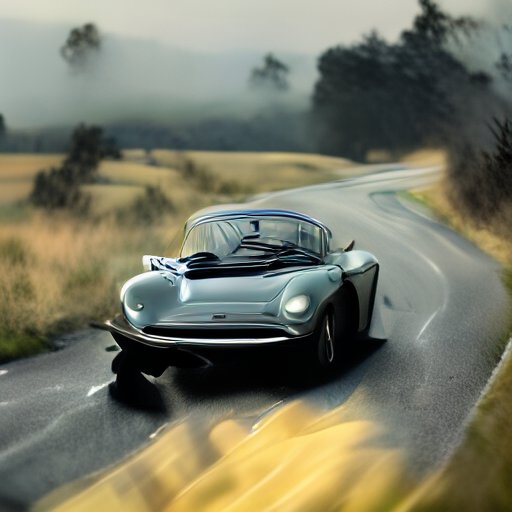}
    \end{minipage}\hspace{0.1cm}%
    \begin{minipage}{0.21\textwidth}
        \includegraphics[width=\linewidth]{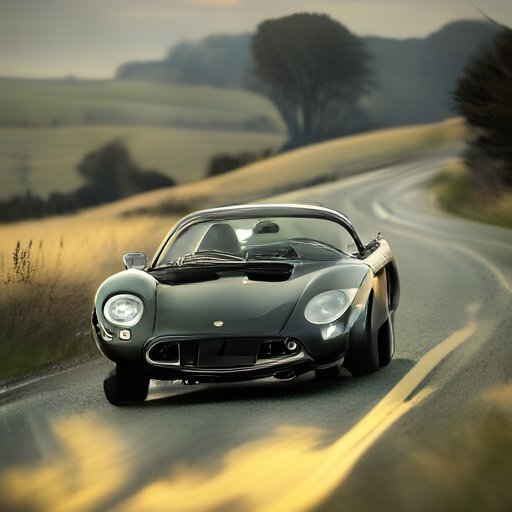}
    \end{minipage}\hspace{0.1cm}%
    \begin{minipage}{0.21\textwidth}
        \includegraphics[width=\linewidth]{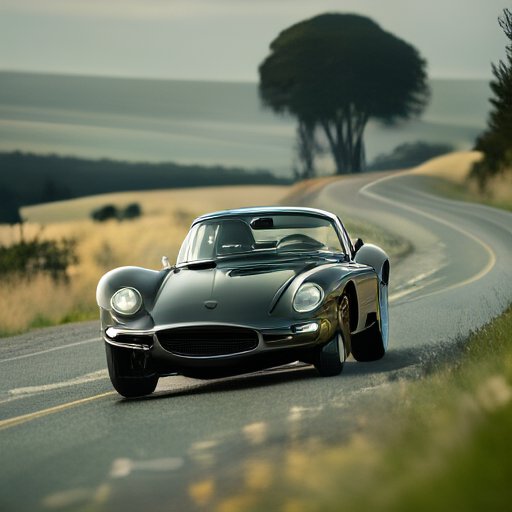}
    \end{minipage}\hspace{0.1cm}%
    \begin{minipage}{0.21\textwidth}
        \includegraphics[width=\linewidth]{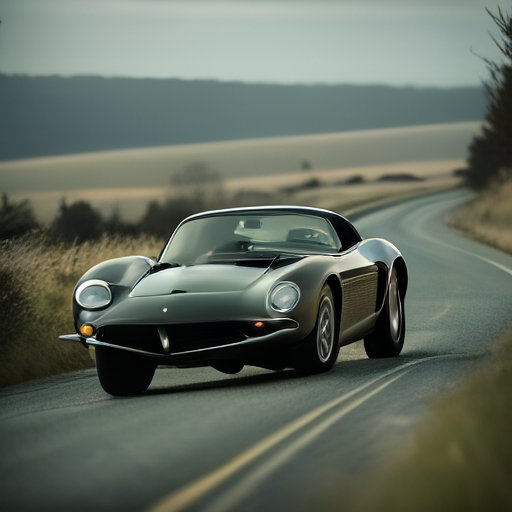}
    \end{minipage}\hspace{0.1cm}%

    \vspace{-0.1cm}

    \begin{minipage}{0.21\textwidth}
        \includegraphics[width=\linewidth]{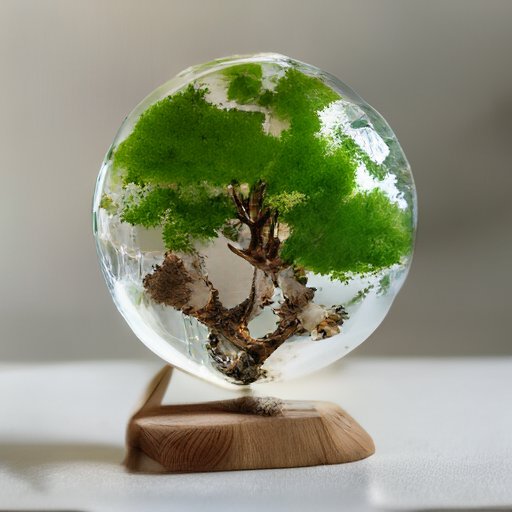}
    \end{minipage}\hspace{0.1cm}%
    \begin{minipage}{0.21\textwidth}
        \includegraphics[width=\linewidth]{graphs/showo512-jpg/t4/test_lmcm_x_photo_12.jpg}
    \end{minipage}\hspace{0.1cm}%
    \begin{minipage}{0.21\textwidth}
        \includegraphics[width=\linewidth]{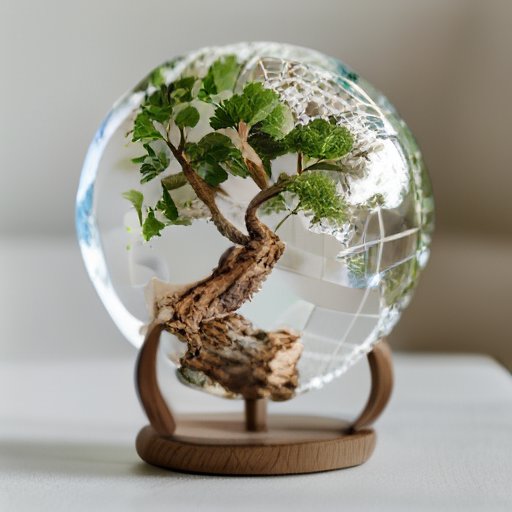}
    \end{minipage}\hspace{0.1cm}%
    \begin{minipage}{0.21\textwidth}
        \includegraphics[width=\linewidth]{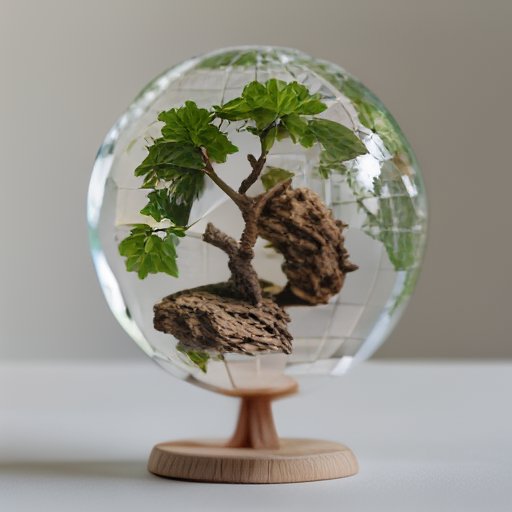}
    \end{minipage}\hspace{0.1cm}%

    \vspace{-0.1cm}

    \begin{minipage}{0.21\textwidth}
        \includegraphics[width=\linewidth]{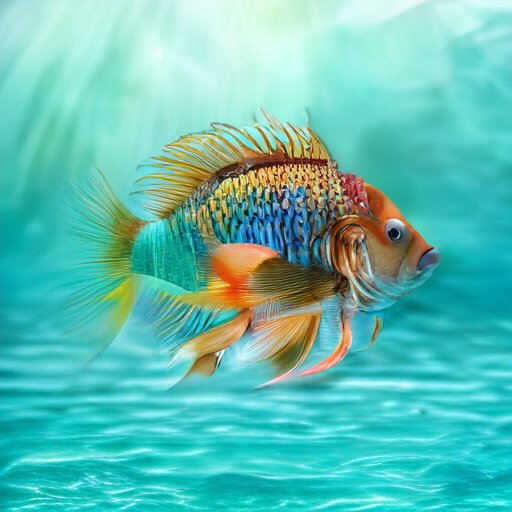}
    \end{minipage}\hspace{0.1cm}%
    \begin{minipage}{0.21\textwidth}
        \includegraphics[width=\linewidth]{graphs/showo512-jpg/t4/test_lmcm_x_photo_7.jpg}
    \end{minipage}\hspace{0.1cm}%
    \begin{minipage}{0.21\textwidth}
        \includegraphics[width=\linewidth]{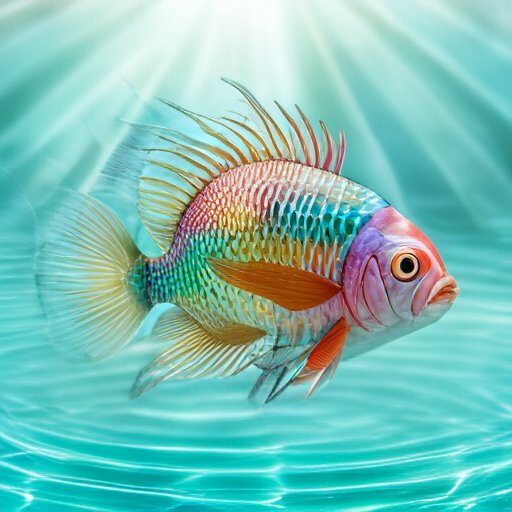}
    \end{minipage}\hspace{0.1cm}%
    \begin{minipage}{0.21\textwidth}
        \includegraphics[width=\linewidth]{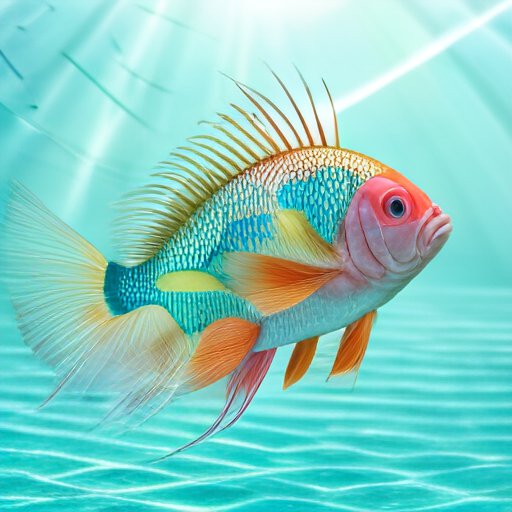}
    \end{minipage}\hspace{0.1cm}%

    \vspace{-0.1cm}

    \begin{minipage}{0.21\textwidth}
        \includegraphics[width=\linewidth]{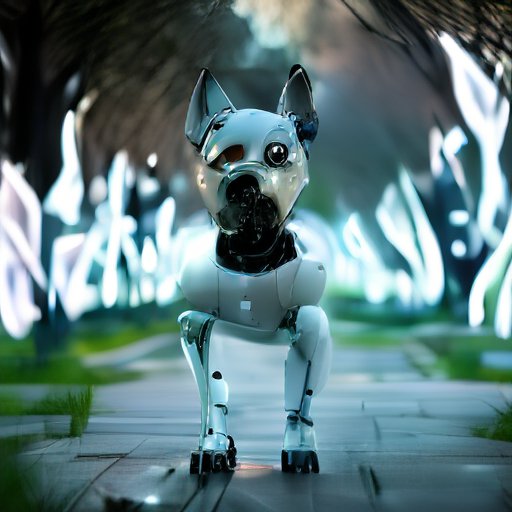}
    \end{minipage}\hspace{0.1cm}%
    \begin{minipage}{0.21\textwidth}
        \includegraphics[width=\linewidth]{graphs/showo512-jpg/t4/test_lmcm_x_photo_19.jpg}
    \end{minipage}\hspace{0.1cm}%
    \begin{minipage}{0.21\textwidth}
        \includegraphics[width=\linewidth]{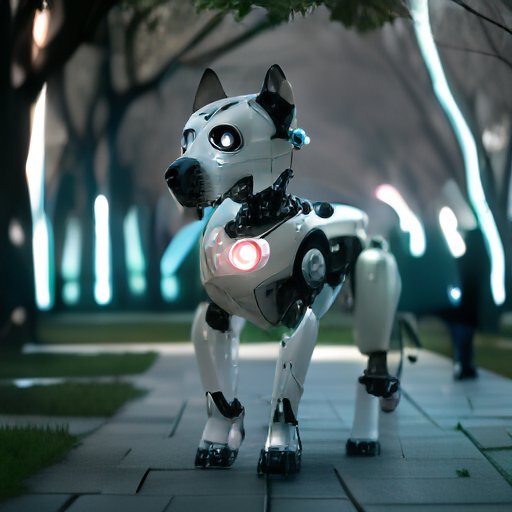}
    \end{minipage}\hspace{0.1cm}%
    \begin{minipage}{0.21\textwidth}
        \includegraphics[width=\linewidth]{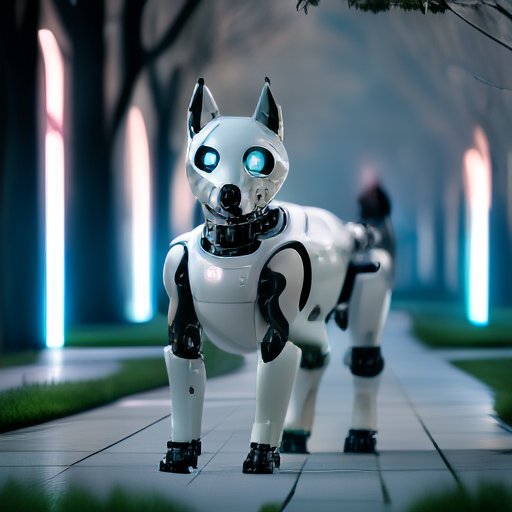}
    \end{minipage}\hspace{0.1cm}%

    \vspace{-0.1cm}

    \begin{minipage}{0.21\textwidth}
        \includegraphics[width=\linewidth]{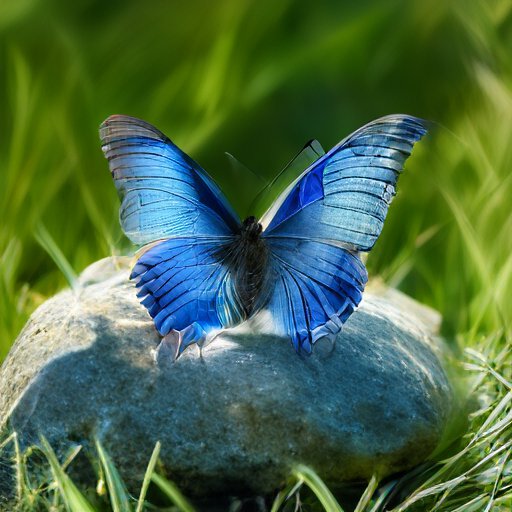}
    \end{minipage}\hspace{0.1cm}%
    \begin{minipage}{0.21\textwidth}
        \includegraphics[width=\linewidth]{graphs/showo512-jpg/t4/test_lmcm_x_photo_10.jpg}
    \end{minipage}\hspace{0.1cm}%
    \begin{minipage}{0.21\textwidth}
        \includegraphics[width=\linewidth]{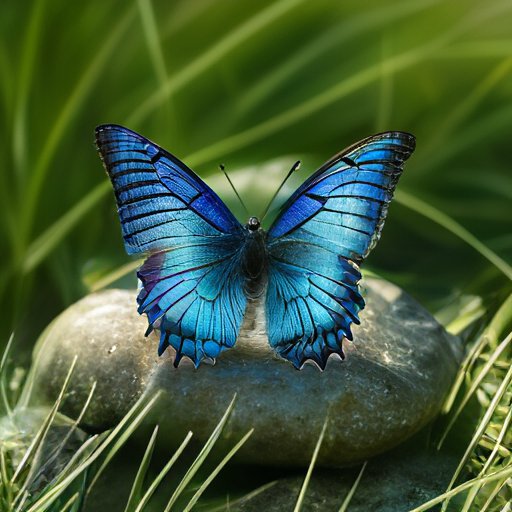}
    \end{minipage}\hspace{0.1cm}%
    \begin{minipage}{0.21\textwidth}
        \includegraphics[width=\linewidth]{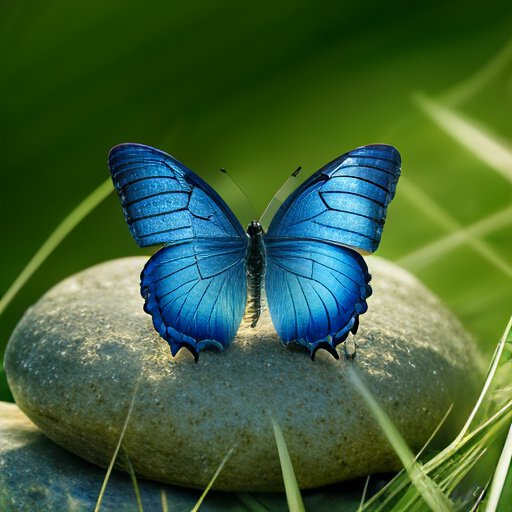}
    \end{minipage}\hspace{0.1cm}%

    \vspace{-0.1cm}

    \begin{minipage}{0.21\textwidth}
        \includegraphics[width=\linewidth]{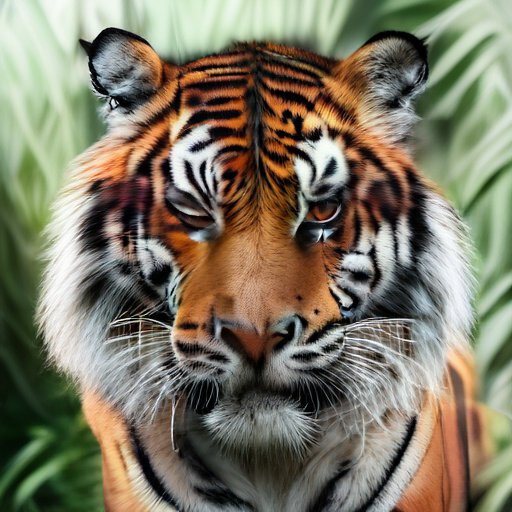}
    \end{minipage}\hspace{0.1cm}%
    \begin{minipage}{0.21\textwidth}
        \includegraphics[width=\linewidth]{graphs/showo512-jpg/t4/test_lmcm_x_photo_32.jpg}
    \end{minipage}\hspace{0.1cm}%
    \begin{minipage}{0.21\textwidth}
        \includegraphics[width=\linewidth]{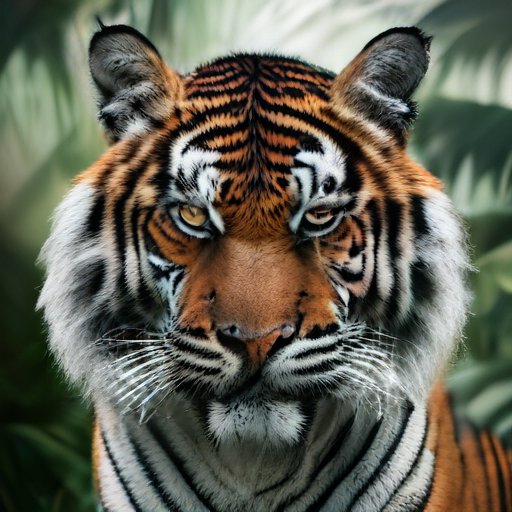}
    \end{minipage}\hspace{0.1cm}%
    \begin{minipage}{0.21\textwidth}
        \includegraphics[width=\linewidth]{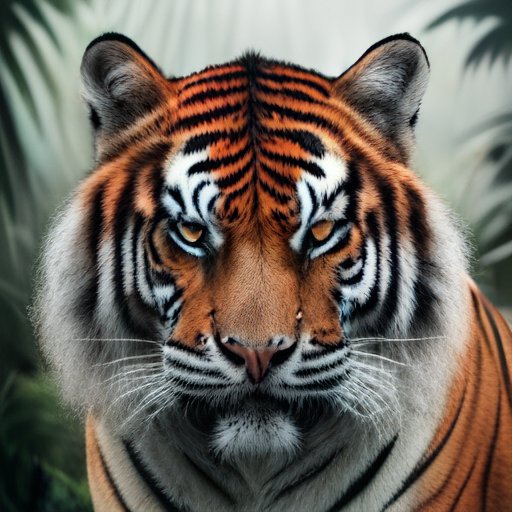}
    \end{minipage}\hspace{0.1cm}%
    \vspace{-0.1cm}

    \caption{\small \textbf{512 $\times$ 512 images generated by UniCMs.} From left to right, the images are generated by UniCMs in 2, 4, 8 and 16 sampling steps without CFG.}
    \label{fig:512display}
\end{figure*}

\begin{figure*}[t]
    \centering

    \begin{minipage}{0.21\textwidth}
        \includegraphics[width=\linewidth]{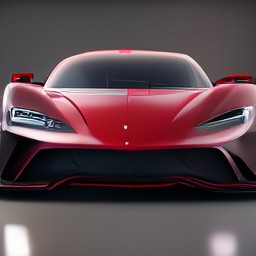}
    \end{minipage}\hspace{0.1cm}%
    \begin{minipage}{0.21\textwidth}
        \includegraphics[width=\linewidth]{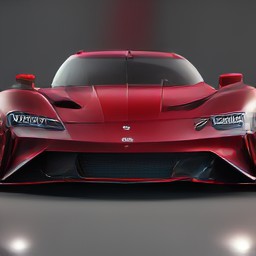}
    \end{minipage}\hspace{0.1cm}%
    \begin{minipage}{0.21\textwidth}
        \includegraphics[width=\linewidth]{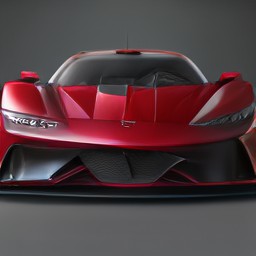}
    \end{minipage}\hspace{0.1cm}%
    \begin{minipage}{0.21\textwidth}
        \includegraphics[width=\linewidth]{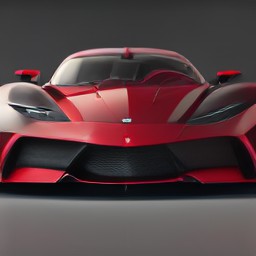}
    \end{minipage}\hspace{0.1cm}%

    \vspace{-0.1cm}

    \begin{minipage}{0.21\textwidth}
        \includegraphics[width=\linewidth]{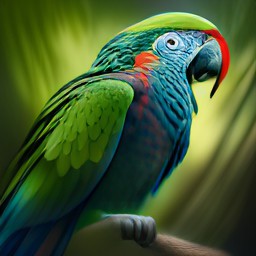}
    \end{minipage}\hspace{0.1cm}%
    \begin{minipage}{0.21\textwidth}
        \includegraphics[width=\linewidth]{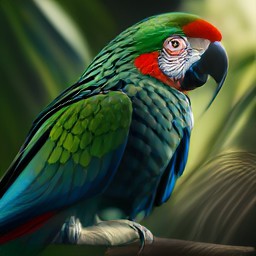}
    \end{minipage}\hspace{0.1cm}%
    \begin{minipage}{0.21\textwidth}
        \includegraphics[width=\linewidth]{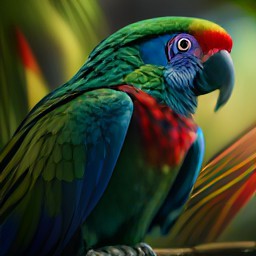}
    \end{minipage}\hspace{0.1cm}%
    \begin{minipage}{0.21\textwidth}
        \includegraphics[width=\linewidth]{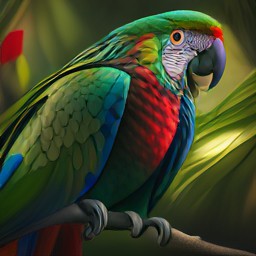}
    \end{minipage}\hspace{0.1cm}%

    \vspace{-0.1cm}

    \begin{minipage}{0.21\textwidth}
        \includegraphics[width=\linewidth]{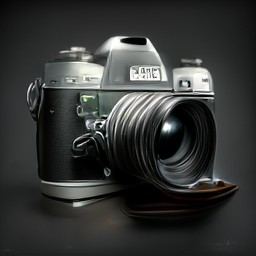}
    \end{minipage}\hspace{0.1cm}%
    \begin{minipage}{0.21\textwidth}
        \includegraphics[width=\linewidth]{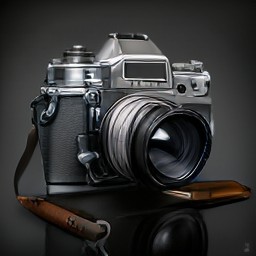}
    \end{minipage}\hspace{0.1cm}%
    \begin{minipage}{0.21\textwidth}
        \includegraphics[width=\linewidth]{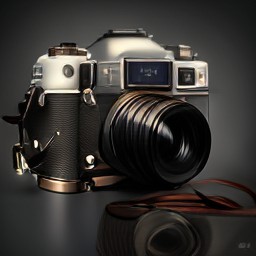}
    \end{minipage}\hspace{0.1cm}%
    \begin{minipage}{0.21\textwidth}
        \includegraphics[width=\linewidth]{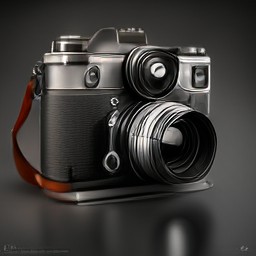}
    \end{minipage}\hspace{0.1cm}%

    \vspace{-0.1cm}

    \begin{minipage}{0.21\textwidth}
        \includegraphics[width=\linewidth]{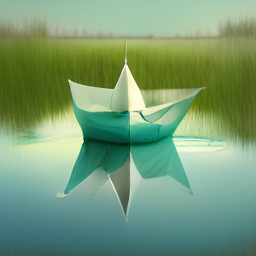}
    \end{minipage}\hspace{0.1cm}%
    \begin{minipage}{0.21\textwidth}
        \includegraphics[width=\linewidth]{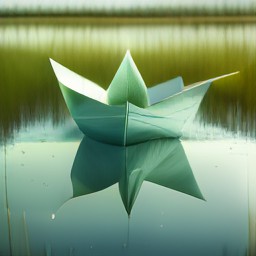}
    \end{minipage}\hspace{0.1cm}%
    \begin{minipage}{0.21\textwidth}
        \includegraphics[width=\linewidth]{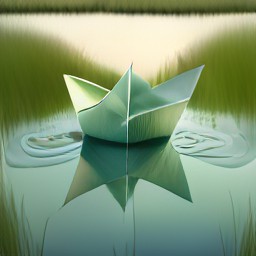}
    \end{minipage}\hspace{0.1cm}%
    \begin{minipage}{0.21\textwidth}
        \includegraphics[width=\linewidth]{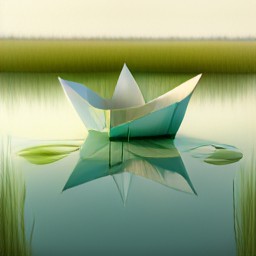}
    \end{minipage}\hspace{0.1cm}%

    \vspace{-0.1cm}

    \begin{minipage}{0.21\textwidth}
        \includegraphics[width=\linewidth]{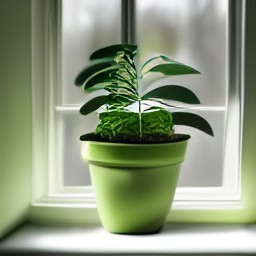}
    \end{minipage}\hspace{0.1cm}%
    \begin{minipage}{0.21\textwidth}
        \includegraphics[width=\linewidth]{graphs/figure8-jpg/pick16/test_lmcm_x_photo_20.jpg}
    \end{minipage}\hspace{0.1cm}%
    \begin{minipage}{0.21\textwidth}
        \includegraphics[width=\linewidth]{graphs/figure8-jpg/pick16/test_lmcm_x_photo_20.jpg}
    \end{minipage}\hspace{0.1cm}%
    \begin{minipage}{0.21\textwidth}
        \includegraphics[width=\linewidth]{graphs/figure8-jpg/pick16/test_lmcm_x_photo_20.jpg}
    \end{minipage}\hspace{0.1cm}%

    \vspace{-0.1cm}

    \begin{minipage}{0.21\textwidth}
        \includegraphics[width=\linewidth]{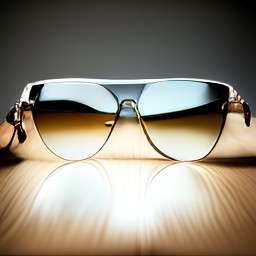}
    \end{minipage}\hspace{0.1cm}%
    \begin{minipage}{0.21\textwidth}
        \includegraphics[width=\linewidth]{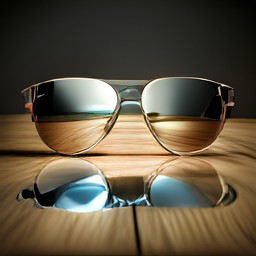}
    \end{minipage}\hspace{0.1cm}%
    \begin{minipage}{0.21\textwidth}
        \includegraphics[width=\linewidth]{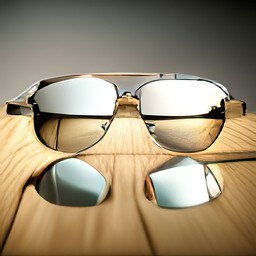}
    \end{minipage}\hspace{0.1cm}%
    \begin{minipage}{0.21\textwidth}
        \includegraphics[width=\linewidth]{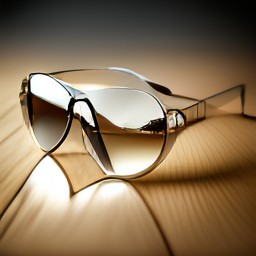}
    \end{minipage}%
    \vspace{-0.1cm}

    \caption{\small \textbf{256 $\times$ 256 images generated by UniCMs.} From left to right, the images are generated by UniCMs in 2, 4, 8 and 16 sampling steps without CFG.}
    \label{fig:display}
\end{figure*}

\section{Additional Image Results}

Figure~\ref{fig:512display} and Figure~\ref{fig:display} show the image generation results for 512 and 256 resolutions respectively. 
UniCMs can generate high-quality images with rich details using only 2 to 4 sampling steps and without CFG. 
\clearpage

\end{document}